\RequirePackage{fix-cm}
\documentclass[11pt,letterpaper]{article}

\usepackage[english]{babel}
\usepackage[numbers,sort&compress]{natbib}
\usepackage{embodiedplayground}
\usepackage{longtable}
\DeclareCaptionFont{teaser}{\fontsize{9}{10.5}\selectfont}
\hypersetup{
  pdftitle={Embodied Playground: A Long-Horizon Agentic Robotics Benchmark for Physical, Strategic, and Spatial Reasoning},
  pdfauthor={Feng Chen, Pei Zhou, Lixuan He, Yunchao Yao, Li Sun, Shenghua Gao, Mingyu Ding, Kaifeng Zhang, Kaichen Zhou, Yi Ma},
  pdfsubject={Technical Report},
  pdfkeywords={robotics, foundation models, embodied agents, benchmark}
}
\reportshorttitle{Embodied Playground}
\reporttype{Technical Report}
\reportversion{Preprint}

\title{\epbrand}
\reportsubtitle{A Long-Horizon Agentic Robotics Benchmark\\
for Physical, Strategic, and Spatial Reasoning}
\author{%
  Feng Chen\textsuperscript{1,2,7,$\dagger$,*}\quad
  Pei Zhou\textsuperscript{1,2,*}\quad
  Lixuan He\textsuperscript{3,4,*}\quad
  Yunchao Yao\textsuperscript{2,5,*}\quad
  Li Sun\textsuperscript{1,*}\\[7pt]
  Shenghua Gao\textsuperscript{1}\quad
  Mingyu Ding\textsuperscript{5}\quad
  Kaifeng Zhang\textsuperscript{2}\quad
  Kaichen Zhou\textsuperscript{4,6}\quad
  Yi Ma\textsuperscript{1,7}%
}
\reportaffiliation{%
  \textsuperscript{1}HKU\quad
  \textsuperscript{2}Sharpa\quad
  \textsuperscript{3}Tsinghua University\quad
  \textsuperscript{4}Stanford University\\
  \textsuperscript{5}University of North Carolina at Chapel Hill\quad
  \textsuperscript{6}MIT\quad
  \textsuperscript{7}Transcengram\\[3pt]
  \textsuperscript{$\dagger$}Project Leader\quad
  \textsuperscript{*}Core Contributor%
}
\date{}

\begin{document}
\newgeometry{left=1in,right=1in,top=.4in,bottom=.65in,footskip=20pt}
\begingroup
\setlength{\parskip}{1pt}
\linespread{1}\selectfont
\makeatletter
\thispagestyle{epfirst}
\noindent\epmark\hspace{8pt}%
{\ttfamily\fontsize{8}{10}\selectfont\color{EPMuted}EMBODIED.\textcolor{EPGreen}{PLAYGROUND}}
\hfill{\ttfamily\fontsize{8}{10}\selectfont\color{EPTeal}\MakeUppercase{\ep@type}}\par
\vspace{2pt}
{\sffamily\bfseries\fontsize{28}{30}\selectfont\raggedright\@title\par}
\vspace{3pt}
{\sffamily\fontsize{12.5}{15}\selectfont\color{EPMuted}\raggedright\ep@subtitle\par}
\vspace{3pt}
\patchcmd{\@author}{\\[7pt]}{\\[2pt]}{}{}
{\sffamily\fontsize{9}{11}\selectfont\@author\par}
\vspace{2pt}
\patchcmd{\ep@affiliation}{\\[3pt]}{\\[1pt]}{}{}
{\sffamily\fontsize{8.3}{10}\selectfont\color{EPMuted}\ep@affiliation\par}
\vspace{2pt}
\noindent\rlap{\textcolor{EPLine}{\rule{\linewidth}{.5pt}}}%
\textcolor{EPGreen}{\rule{42pt}{2pt}}\par
\makeatother
\renewenvironment{abstract}{%
  \begin{tcolorbox}[epbase,breakable=false,left=8pt,right=8pt,top=5pt,bottom=5pt,
    before skip=3pt,after skip=5pt,
    fonttitle=\sffamily\bfseries\fontsize{8.5}{10}\selectfont,
    attach title to upper={\par\vspace{2pt}} ,title={ABSTRACT}]
  \fontsize{9.3}{11.2}\selectfont\ignorespaces
}{\end{tcolorbox}}
\begin{abstract}
Foundation models offer a promising basis for robotic agents, but long-horizon tasks require them to sustain coherent reasoning and reliable physical execution across interdependent stages. We introduce \textbf{Embodied Playground}, a benchmark for evaluating how foundation-model agents coordinate physical, strategic, and spatial reasoning throughout such tasks. Agents must track task progress, satisfy intermediate constraints, and revise their plans using execution feedback as earlier actions shape subsequent decisions. The playground comprises 410 task environments spanning basic manipulation, sequential puzzles, and multi-stage household activities, with variations in planning depth, geometric constraints, and physical execution demands. We curate \textbf{EP-100}, a representative selection of 100 tasks across five suites and 16 task types, together with the EP-Easy and EP-Hard difficulty splits and the compact EP-20 selection. A shared protocol specifies observation and tool access, resource budgets, repeated trials, and independent checks of goal completion, task-rule compliance, and physical validity. Embodied Playground provides a unified framework for studying whether foundation-model agents can sustain effective reasoning and action over extended task horizons.
The project website is available at
\href{https://winniechen2002.github.io/Agent-bench-website/}{Embodied Playground Website}.
\end{abstract}

\noindent\begin{minipage}{\linewidth}
  \centering
  \includegraphics[width=\linewidth]{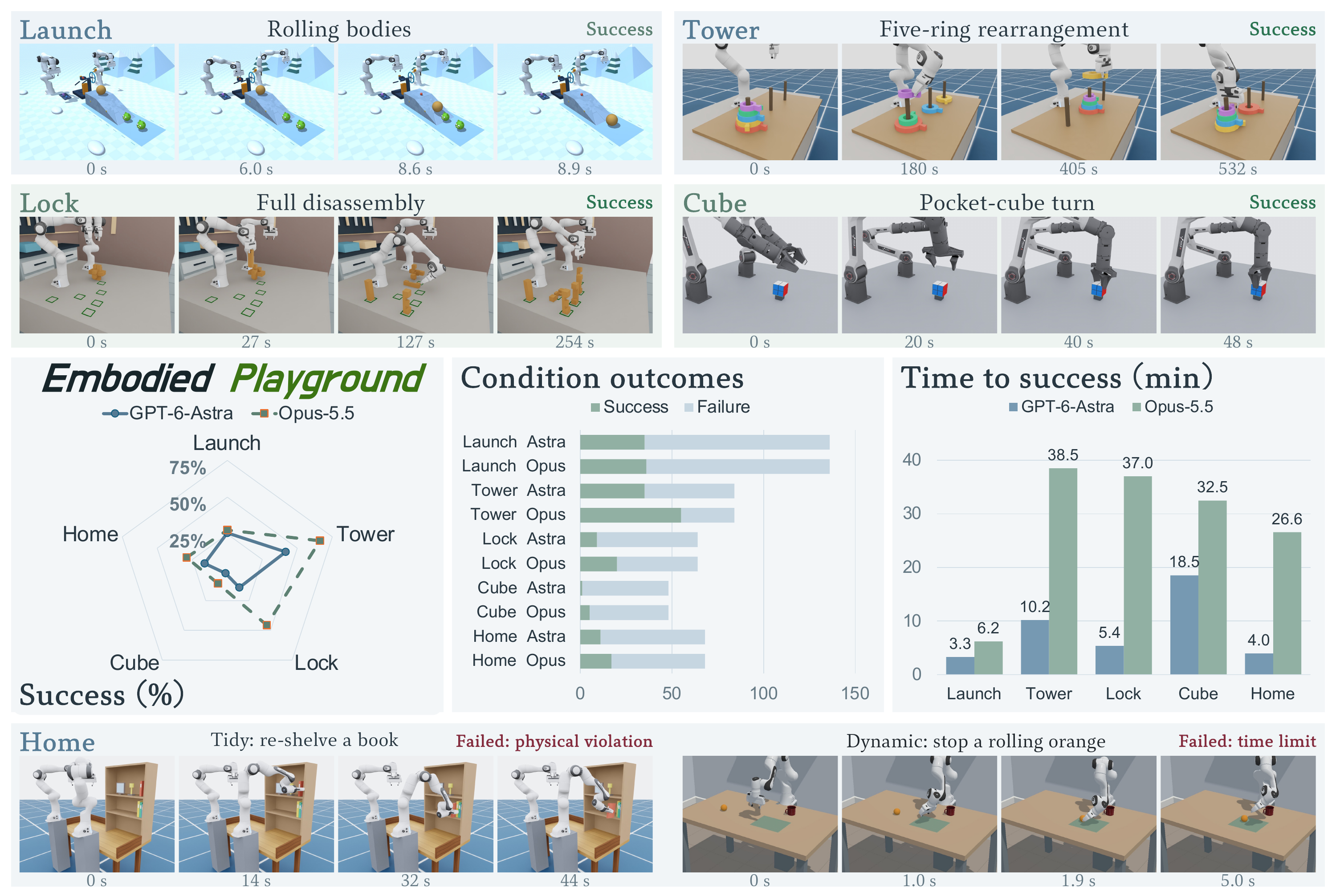}
  \captionsetup{font={teaser,sf},skip=4pt,hypcap=false}
  \captionof{figure}{\textbf{Embodied Playground at a glance.} Six trajectories illustrate Launch, Tower, Lock, Cube, and two Home environments, including successes and failures. The middle row compares GPT-6-Astra and Opus-5.5-extra by suite on Pass@2, condition outcomes, and mean duration of successful attempts. Frame timestamps use simulated seconds; the timing chart uses wall-clock minutes.}
  \label{fig:teaser}
\end{minipage}\par
\endgroup
\restoregeometry

\section{Introduction}
\label{sec:introduction}
Foundation models have shown promising capabilities in robotic tasks, supported by their ability to interpret visual observations, reason about spatial relationships and physical interactions, and formulate strategies. \citep{anthropic2026claudefable51,mirros2026sspace,openai2026gpt6astra, anthropic2026claudeopus55} Existing robotics benchmarks largely emphasize the performance and generalization of control policies. \citep{liu2023liberobenchmarkingknowledgetransfer,zhou2026liberoprorobustfairevaluation, zhang2026liberomaxrobotpoliciesadapt, chen2026dexholdemagenticroboticsbenchmark} This leaves a complementary question: can a foundation-model agent sustain these capabilities over long-horizon robotic tasks, where actions must establish the conditions for later stages and execution errors can affect subsequent decisions? Addressing this question requires evaluating how an agent tracks task progress, satisfies intermediate constraints, and adapts its plan throughout physical execution.

We introduce \textbf{Embodied Playground}, a benchmark for studying long-horizon robotic problem solving by foundation-model agents. Tasks couple physical, strategic, and spatial reasoning across dependent stages: an agent must determine what to do next, execute it through the robot interface, and use the resulting observations to update its understanding of the remaining task. Earlier actions can change object configurations, consume resources, or constrain later choices, making sustained coordination between planning and execution central to successful completion. Given a task description, permitted observations, and a specified set of tools, the agent must ground its decisions in the observed scene and revise its plan using execution feedback. Success requires achieving the final goal while respecting task rules and physical constraints throughout the episode.

Embodied Playground comprises 410 robotic task environments, ranging from basic object manipulation to long-horizon planning and dexterous coordination. These tasks vary in planning depth, geometric constraints, and physical execution demands. Tower of Hanoi requires executing legal move sequences while preserving valid intermediate configurations; cube puzzles require keeping track of puzzle states and object orientations through successive physical turns; and Kongming locks couple assembly or disassembly order with geometric feasibility and contact-rich manipulation. Multi-stage household tasks combine ordered interactions with object-placement and final-scene requirements. Launcher tasks connect trajectory prediction and causal effects with finite ammunition and, in selected scenes, target-order constraints. Shorter manipulation, precision, and dynamics tasks complement these extended sequences by probing constituent capabilities across multiple robot embodiments.

To facilitate evaluation, we curate \textbf{EP-100}, a representative selection of 100 tasks organized into five subsets: \emph{Launch}, \emph{Tower}, \emph{Lock}, \emph{Cube}, and \emph{Home}. Together, these subsets cover 16 task types and provide focused evaluations alongside the full playground. EP-100 is divided into \textbf{EP-Easy} and \textbf{EP-Hard} by ranking the tasks of each type according to their workload, and \textbf{EP-20} offers a compact selection that spans both splits. All tasks follow a common interaction interface and evaluation protocol, with specified observation and tool access, execution budgets, repeated trials, and task-specific physical success criteria.

Our contributions are threefold:
\begin{itemize}
    \item \textbf{An agent-centered perspective on long-horizon robotic problem solving.} We evaluate a foundation-model agent's ability to sustain coordinated perception, reasoning, tool use, and action across interdependent task stages, using execution feedback to track progress, revise plans, and satisfy task constraints throughout an episode.
    \item \textbf{A diverse playground spanning task horizons and execution demands.} We introduce 410 robotic task environments spanning basic manipulation and multi-stage problem solving, with variations in planning depth, geometry, and physical execution requirements. EP-100 provides a representative selection across five suites and 16 task types, complemented by two difficulty splits and the compact EP-20 selection.
    \item \textbf{An episode-level protocol for evaluating completion and validity.} We specify observation and tool-access levels, resource budgets, and repeated-trial conditions, with independent checks of final goal completion, task-rule compliance, and physical validity over the executed trajectory.
\end{itemize}

\section{Related Work}
\label{sec:related-work}

\subsection{Foundation-model agents and harnesses for Robotics.}
Early systems such as SayCan, Code as Policies, and ProgPrompt connected language-model reasoning
to robot action through skill selection or executable
programs~\cite{ahn2022saycan,liang2023codeaspolicies,singh2023progprompt}.
Inner Monologue and subsequent systems such as REFLECT and Statler used execution feedback,
failure explanations, or maintained world-state estimates to inform later
actions~\cite{lit2022innermonologue,lit2023reflect,lit2023statler}.
Spatial grounding developed through representations such as scene graphs, value maps, and
keypoint constraints~\cite{lit2023sayplan,lit2023voxposer,lit2024rekep}.
Systems developed in 2024--2025, such as RoboScript, ROS-LLM, and RAI, connected model-generated
decisions to robot software and execution services~\cite{lit2024roboscript,lit2024rosllm,lit2025rai}.
Hi Robot, Agentic Robot, and Code-as-Monitor illustrate related directions in hierarchical action
selection, execution coordination, and failure monitoring~\cite{shi2025hirobot,lit2025agenticrobot,zhou2025codeasmonitor}.
The interface between reasoning and execution has become an explicit design focus, exemplified by
Robot Context Protocol and later visual, semantic, and native-software interfaces in VIA,
Show-Harness, and OpenRUA~\cite{lit2025robotcontextprotocol,lit2026via,lit2026showharness,lit2026openrua}.
Recent tool-based harnesses also examine perception support, context management, and outcome
verification~\cite{lit2026guava,lit2026thea,lit2026etaopeneta,lit2026regenharness,lit2026roboharnessk1,lit2026robotuse}.
Alongside these interface designs, systems such as RoboHarness, RATs, ASPIRE, and CarryOn use memory
or reusable skills to adapt across interactions~\cite{lit2026roboharnessmemoryvla,lit2026rats,lit2026aspire,lit2026carryon}.
We offer a common basis for assessing these diverse agent designs through their physical outcomes. Its protocol makes differences in assistance and resource use explicit, supporting fair comparisons of task completion, consistency, and efficiency.

\subsection{Simulation infrastructure and robotic policy benchmarks.}
Simulation platforms and task benchmarks serve complementary roles in robotic policy evaluation.
Early simulators such as Gazebo and MuJoCo, followed by platforms such as PyBullet, SAPIEN, and
Isaac Gym, established infrastructure for robotic
experiments~\cite{koenig2004gazebo,todorov2012mujoco,coumans2021pybullet,xiang2020sapien,makoviychuk2021isaacgym}.
Building on simulator backends, frameworks such as robosuite and ORBIT package environments and
robot-learning workflows, with more recent examples including Isaac Lab, MuJoCo Playground, and
ManiSkill3~\cite{zhu2020robosuite,mittal2023orbit,nvidia2025isaaclabgpuacceleratedsimulation,zakka2025mujocoplayground,tao2025maniskill3}.
Task benchmarks specify the learning problems and evaluation protocols.
Early benchmarks such as Meta-World examine multi-task and meta-reinforcement learning, while
RLBench provides manipulation tasks with demonstrations~\cite{yu2020metaworld,james2019rlbench}.
Subsequent suites study generalizable manipulation, language-conditioned action sequences, and
lifelong transfer, as illustrated by ManiSkill, CALVIN, ManiSkill2, and
LIBERO~\cite{mu2021maniskillgeneralizablemanipulationskill,mees2022calvin,gu2023maniskill2,liu2023libero}.
Household activities and longer manipulation tasks feature in benchmarks such as BEHAVIOR-1K,
RoboCasa, VLABench, and RoboCasa365~\cite{li2023behavior1k,nasiriany2024robocasa,zhang2025vlabench,nasiriany2026robocasa365}.
Recent evaluation also examines robustness to variation, for example in Colosseum, RoboTwin~2.0,
LIBERO-Plus, RoboLab, and Colosseum~V2~\cite{pumacay2024colosseum,chen2025robotwin2,fei2026liberoplus,yang2026robolab,morgan2026colosseum2}.
SIMPLER and RoboDojo connect policy evaluation in simulation with real-robot
performance~\cite{li2024simpler,chen2026robodojo}.
Embodied Playground adopts this foundation and organizes its evaluation around a foundation-model agent's problem-solving episode. Its combination of abstract challenges and household tasks makes reasoning demands central to evaluation while preserving the need for successful physical execution.

\subsection{Embodied-agent benchmarks.}
Embodied-agent benchmarks examine reasoning through task performance under specified observation
and action interfaces.
Earlier examples examined physical puzzles in PHYRE and language-guided robot planning in
LoHoRavens and RoboEval~\cite{bakhtin2019phyre,zhang2023lohoravens,lit2023codebotlerroboeval}.
Benchmarks such as LoTa-Bench and Embodied Agent Interface subsequently evaluated task planning and
its component capabilities~\cite{choi2024lotabench,lit2024embodiedagentinterface}.
In 2025, EmbodiedEval and EmbodiedBench evaluated multimodal agents in simulated tasks, while PARTNR
examined collaborative household planning~\cite{lit2025embodiedeval,yang2025embodiedbench,lit2024partnr}.
Physical reasoning benchmarks such as PhyBlock and DeepPHY, followed by COIN and KinDER, probe
assembly planning, physical prediction, causal interaction, and kinematic or dynamic
constraints~\cite{lit2025phyblock,lit2025deepphy,wang2026coin,lit2026kinder}.
Recent benchmarks examine the interaction interface itself: CaP-X and Embodied Skill Bench vary
control abstractions and supplied capabilities, while VABench studies active perception and metric
control~\cite{lit2026capxcapbench,han2026embodiedskillbench,zhang2026vabench}.
CodeActionBench and LIBERO-Agent evaluate direct robot interaction, while Embodied Agent Arena
connects component capabilities to task completion~\cite{lit2026codeactionbench,lit2026liberoagent,lit2026embodiedagentarena}.
Longer execution and broader task coverage feature in benchmarks such as EmbodiedSWE and
RobotWorld~\cite{lit2026embodiedswe,lit2026robotworld}.
LADDER varies the number of subgoals within the same environment~\cite{petkov2026ladder}.
SCOPER studies recovery under controlled skill failures, while SEAR isolates the contribution of
retained experience under controlled interaction budgets~\cite{lit2026scoper,lit2026sear}.
Physical-robot evaluations also appear in Words to World and the AA-Robotics research
preview~\cite{lit2026wordstoworld,artificialanalysis2026aarobotics}.
Our structured task coverage supports interpretable comparisons across reasoning demands and robot embodiments. It allows performance to be examined as task demands increase, helping identify strengths and limitations that a single aggregate score may obscure.

\section{Embodied Playground}
\label{sec:benchmark}

Embodied Playground places foundation-model agents in simulated robotic environments where task completion requires both reasoning and physical execution. Each environment specifies a goal, an initial scene, a robot embodiment, and rules governing valid interaction. The collection combines familiar puzzles and everyday manipulation with controlled variations in geometry, dynamics, and planning depth.

\subsection{Agent--Environment Interaction}
\label{sec:interaction}
\begin{figure}[htbp]
\centering
\includegraphics[width=\linewidth]{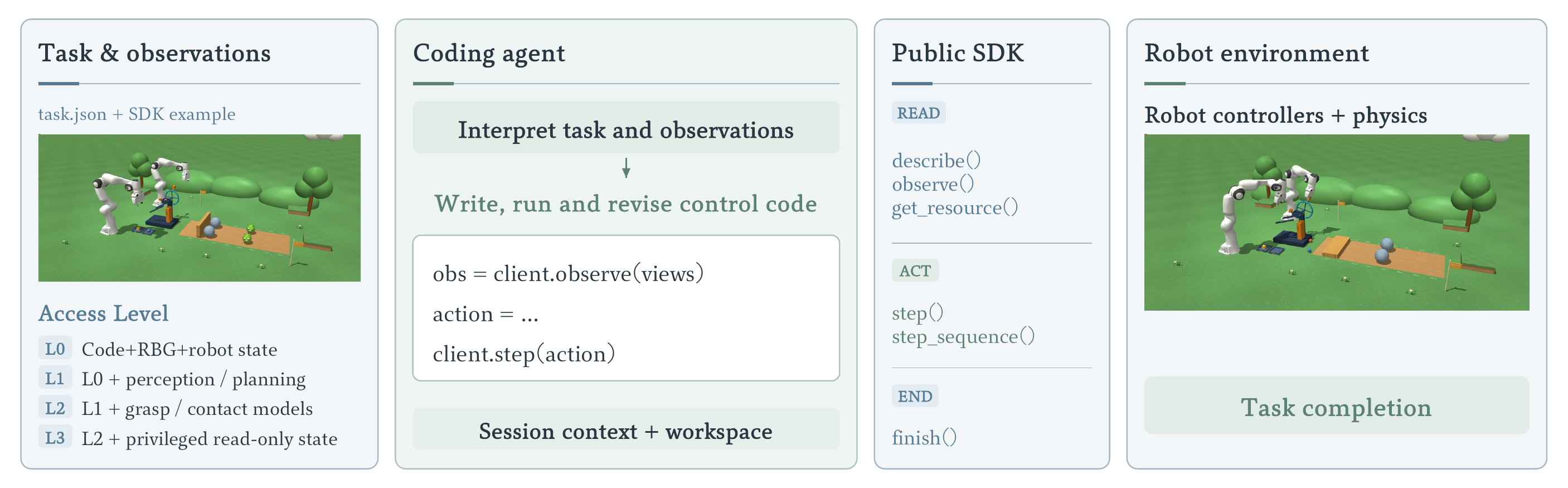}
\caption{Agent--environment interaction in Embodied Playground. The agent writes and executes code through the public SDK and revises its actions using environment feedback. Access levels provide cumulative capabilities. Launcher images show an illustrative reference-controller rollout.}
\label{fig:agent-pipeline}
\end{figure}

An agent receives a task description, the robot's control specification, and observations allowed by the evaluation setting. Public observations include camera images, robot proprioception, and declared task metadata; additional state information is governed by the selected access level. The agent uses these inputs to interpret the scene, construct a plan, and produce robot commands, optionally invoking permitted perception or planning tools. Subsequent observations provide feedback for tracking progress, correcting execution errors, and revising the plan. Figure~\ref{fig:agent-pipeline} summarizes this interaction.

All environments share four core operations: \mbox{\texttt{describe}} exposes task and interface specifications; \mbox{\texttt{observe}} returns permitted observations; \mbox{\texttt{step}} applies an action and advances simulation; and \mbox{\texttt{finish}} ends the episode. The common interface preserves embodiment-specific action spaces, such as arm joint targets, gripper commands, and finger or floating-hand controls. Observation reads leave physical time unchanged, and initialization and reset are managed by the environment. Actions take effect through the declared robot controllers and task mechanisms. An independent evaluator checks goal completion, task-rule compliance, and physical validity, including any required release and stability conditions. This interaction loop connects high-level decisions to their physical consequences; access levels and scoring are detailed in Section~\ref{sec:protocol}.

\subsection{Task Environments and Robot Embodiments}
\label{sec:environments}
Table~\ref{tab:playground} summarizes the 410 task environments in the playground. Each entry pairs a task configuration with a robot embodiment. For example, 40 Hanoi configurations are instantiated with three robot embodiments, yielding 120 entries. Seeds and initial operating modes specify trial conditions within an entry. The environments use MuJoCo \citep{todorov2012mujoco} for launcher tasks, Isaac Lab\citep{nvidia2025isaaclabgpuacceleratedsimulation} for the puzzle and bookshelf families, and ManiSkill\citep{mu2021maniskillgeneralizablemanipulationskill} for the additional household, geometry, and dynamics tasks.

\begin{table}[htbp]
\centering
\caption{The full Embodied Playground collection. Counts denote task--embodiment combinations. The ManiSkill group contains ten tasks each for long-horizon manipulation, geometry, and dynamics.}
\label{tab:playground}
\small
\begin{tabularx}{\linewidth}{@{}l r Y Y@{}}
\toprule
\eptablehead
\textbf{Task family} & \textbf{Count} & \textbf{Task focus} & \textbf{Embodiments} \\
\midrule
Launcher & 100 & Aiming, indirect impacts, fracture, target constraints & Dual Panda \\
Tower of Hanoi & 120 & Legal rearrangement, grasping, peg insertion & Panda; FR3--Sharpa; floating Sharpa \\
Pocket cube & 20 & Prescribed layer turns and regrasping & ALOHA; dual floating Sharpa \\
Rubik's cube & 60 & Cube solving and physical face turns & Panda; floating Sharpa \\
Kongming lock & 60 & Partial and full assembly and disassembly & Dual Panda; dual floating Sharpa \\
Bookshelf & 20 & Sorting, arrangement, and clutter removal & Dual Panda \\
ManiSkill tasks & 30 & Household sequences, precision, and dynamics & Fetch; Panda; Panda--Sharpa; xArm7--Allegro; UR10e--Shadow \\
\midrule
\textbf{Total} & \textbf{410} & & \\
\bottomrule
\end{tabularx}
\end{table}

Task variations couple discrete goals with continuous control. Hanoi spans shortest legal plans of 1, 3, 9, and 31 moves, with variations in support height, insertion clearance, mass, friction, and obstacles. Pocket-cube tasks require state transitions under a prescribed face-order interlock, whereas Rubik's-cube tasks require restoring uniform face colors using declared auxiliary fixtures for support, grasp attachment, and layer control. Kongming locks range from inserting or removing a key to assembling or disassembling a six-piece structure. Bookshelf tasks combine category-dependent placement, orientation, and packing requirements. The ManiSkill tasks extend this range to activities such as preparing a tea station, inserting a keyed connector, catching a moving ball, and guiding a spring-closing door. Shared task definitions across selected embodiments support comparisons between parallel-gripper manipulation and dexterous-hand control.

\subsection{Spatial, Physical, and Strategic Reasoning}
\label{sec:reasoning}
\textbf{Spatial reasoning} connects visual observations to object configurations and feasible motions. Agents must relate task-relevant positions and orientations to the robot's workspace, account for occlusion, and identify geometric compatibility. Examples include aligning a ring with a peg, orienting a notched lock piece for insertion, tracking cube faces through reorientation, and arranging upright books within a narrow shelf. These tasks require geometric understanding to guide grasp selection, motion, and placement.

\textbf{Physical reasoning} concerns how actions change the scene through forces, contact, and motion. Launcher tasks combine trajectory prediction with indirect effects such as rolling objects, falling structures, and glass fracture. Manipulation tasks introduce mass and friction variations, stability requirements, and dynamic interactions such as stopping a rolling object or controlling a spring-loaded articulation. Agents must connect intended outcomes to executable controls and use observed motion to adjust subsequent actions.

\textbf{Strategic reasoning} concerns action dependencies and decisions over extended horizons. Hanoi requires maintaining legal intermediate states; lock assembly requires an order compatible with the interlocking geometry; and household tasks combine multiple milestones into a final arrangement. Launcher scenes additionally impose finite ammunition, target-order requirements, and protected objects, making later decisions depend on earlier impacts and the remaining scene. These demands interact: a viable strategy must be geometrically feasible and physically executable, while execution feedback can change which strategy is appropriate.

\section{EP-100: A Representative Evaluation Set}
\label{sec:ep100}

\subsection{Task Selection}
\label{sec:ep100-suites}

All EP subsets evaluate launcher tasks only in the \emph{loaded} mode, with the first projectile preloaded. Ground-mode initialization is outside their evaluation scope. Other task families use their default initialization mode. Each task therefore defines one evaluation group with four prescribed conditions; a separate attempt budget determines whether Pass@1 alone or both Pass@1 and Pass@2 are measured.
EP-100 selects 100 of the 410 environments so that a complete evaluation remains affordable and still reflects the breadth of the playground. The selection favors tasks that most agents can attempt through the shared runtime. Parallel-gripper embodiments therefore receive most of the slots, and each dexterous-hand family contributes a smaller sample drawn from its easier native configurations. Within every family the selected tasks follow the family's own difficulty axis, such as the launcher tier, the Hanoi planning depth, or the subtask and pose of a lock, so that the native difficulty range of each family remains visible in the selection. Launcher scenes are drawn from the 80 scenes with a recorded full-task success in the task atlas, which establishes that each selected scene can be completed within its shot budget. Several case identifiers recur across embodiments, and this shared structure allows the same task definition to be compared between parallel grippers and dexterous hands.

The selected tasks form five suites, and each suite divides into task types that share a single dominant demand (Table~\ref{tab:ep100-suites}). Launch progresses from open-field aiming to scenes whose outcome depends on indirect causal chains, on the fracture of glass panes, or on rules that protect or order the targets. Tower varies the planning depth and the physical conditions of the Hanoi board and then repeats representative boards with dexterous hands. Lock covers assembly and disassembly of the Kongming lock with two Panda arms and with two floating Sharpa hands, and Cube combines Rubik's-cube solving with prescribed pocket-cube turns. Home extends the evaluation to household settings, where its bookshelf and long-horizon types test how an agent structures a multi-step task and its precision and dynamics types test how accurately it controls contact and motion. Because each type isolates one dominant demand, results aggregated by type indicate which capability limits an agent.

\begin{table}[htbp]
\centering
\caption{EP-100 suites and task types. E and H give the number of tasks of each type in EP-Easy and EP-Hard.}
\label{tab:ep100-suites}
\small
\begin{tabularx}{\linewidth}{@{}l l r r Y@{}}
\toprule
\eptablehead
\textbf{Suite} & \textbf{Type} & \textbf{Tasks} & \textbf{E\,/\,H} & \textbf{Principal demand} \\
\midrule
Launch & Aim & 7 & 4\,/\,3 & Direct shots at single or paired targets, including azimuth aiming \\
 & Causal & 13 & 6\,/\,7 & Indirect hits through rolling bodies, dominoes, and collapsing structures \\
 & Glass & 5 & 3\,/\,2 & Breakable glass panes with fracture thresholds \\
 & Rules & 9 & 4\,/\,5 & Ordered or protected targets and recovery after a miss \\
\midrule
Tower & Plan & 5 & 3\,/\,2 & Planning depth from 1 to 31 moves on a standard board \\
 & Physics & 8 & 4\,/\,4 & Physical board variations under the same legal-move rule \\
 & Dex & 8 & 4\,/\,4 & Representative boards with an FR3--Sharpa or floating Sharpa hand \\
\midrule
Lock & Assemble & 6 & 3\,/\,3 & Inserting the key, the key and bar, or the full lock with two Panda arms \\
 & Disassemble & 6 & 3\,/\,3 & Removing the key, the key and bar, or all six pieces with two Panda arms \\
 & Dex & 4 & 2\,/\,2 & Assembly and disassembly with two floating Sharpa hands \\
\midrule
Cube & Rubik & 8 & 4\,/\,4 & Solving scrambles of 5, 10, or 15 tokens through the auxiliary fixture \\
 & Pocket & 4 & 2\,/\,2 & Prescribed layer turns with ALOHA grippers or two floating Sharpa hands \\
\midrule
Home & Tidy & 5 & 2\,/\,3 & Organizing a bookshelf and correcting misplaced objects \\
 & Long & 4 & 2\,/\,2 & Multi-step household sequences with a mobile Fetch robot \\
 & Precise & 4 & 2\,/\,2 & Insertion, hooking, and narrow-gap placement with a Panda arm \\
 & Dynamic & 4 & 2\,/\,2 & Intercepting, pushing, and closing objects in motion with a Panda arm \\
\midrule
\textbf{Total} & 16 types & \textbf{100} & 50\,/\,50 & \\
\bottomrule
\end{tabularx}
\end{table}

\subsection{Evaluation Subsets}
\label{sec:ep100-difficulty}
EP-100 is divided into two disjoint splits of 50 tasks, EP-Easy and EP-Hard. The assignment is derived from the task definitions, which keeps each split fixed as agents improve and makes it independent of any particular evaluation outcome. Each task receives a workload score from the quantities that make its family demanding, such as the move count of a Hanoi board, the scramble length of a cube, or the subtask and pose of a lock, and the score rises for dexterous-hand embodiments. For Launch, only the loaded-mode difficulty contributes to the workload score; ground-mode pickup requirements are excluded. The tasks of each type are ranked by this score, and the lower half of every type forms EP-Easy. A type with an odd number of tasks places its median task in the half whose neighboring score is closer, and the remaining ties are resolved so that each split holds exactly 50 tasks and the two splits differ by at most one task within any suite. Both splits therefore contain all 16 task types in nearly the same proportions, so a gap between EP-Easy and EP-Hard results reflects difficulty within a comparable task mix. Appendix~\ref{app:ep100} defines the score for each suite.

EP-20 is a compact selection for rapid and representative evaluation. It takes ten tasks from EP-Easy and ten from EP-Hard, which together cover every task type and robot embodiment of EP-100 on all three simulators, and its six launcher scenes span the five native tiers. Half of its Hard tasks have been completed at least once in pilot agent runs on the same or closely related scenes, so EP-20 can register progress on difficult tasks before an agent masters the full set. Every subset is scored with the suite-balanced aggregation of Section~\ref{sec:metrics}, which gives each of the five suites one fifth of the subset score. Table~\ref{tab:ep-splits} summarizes the four subsets, and Appendices~\ref{app:ep100}--\ref{app:ep20} list their members.

\begin{table}[htbp]
\centering
\caption{EP-100 subsets. Counts give tasks per suite and registered conditions per fixed evaluation setting, with loaded-only launcher initialization. Each condition uses one attempt for Pass@1 and at most two for Pass@2. EP-Easy and EP-Hard partition EP-100, and EP-20 draws ten tasks from each.}
\label{tab:ep-splits}
\small
\begin{tabularx}{\linewidth}{@{}Y r r r r r r r@{}}
\toprule
\eptablehead
\textbf{Subset} & \textbf{Tasks} & \textbf{Launch} & \textbf{Tower} & \textbf{Lock} & \textbf{Cube} & \textbf{Home} & \textbf{Conditions} \\
\midrule
EP-100 & 100 & 34 & 21 & 16 & 12 & 17 & 400 \\
EP-Easy & 50 & 17 & 11 & 8 & 6 & 8 & 200 \\
EP-Hard & 50 & 17 & 10 & 8 & 6 & 9 & 200 \\
EP-20 & 20 & 6 & 4 & 3 & 3 & 4 & 80 \\
\bottomrule
\end{tabularx}
\end{table}

\section{Evaluation Protocol}
\label{sec:protocol}

Evaluation uses four prescribed trial conditions per task and four levels of observation and tool access. The paper adopts loaded-only launcher initialization.\footnote{Protocol: \nolinkurl{environment-four-trial-v2}; permissions: \nolinkurl{2026-10-09-environment-l0-l3}.} The protocol fixes the information and tools available to an agent, the conditions of repeated interaction, and the physical criteria for success. All trials use the shared environment runtime introduced in Section~\ref{sec:interaction}.

\subsection{Access Levels}
\label{sec:access}
Capability levels specify available assistance independently of task difficulty. The four levels are cumulative:
\begin{itemize}
    \item \textbf{L0: code-enabled control.} The agent may use Python's standard library, NumPy, Pillow, and self-written algorithms, including its own inverse kinematics, dynamics prediction, and planning. Inputs comprise RGB images, proprioception, camera calibration, and public task and robot information. Prebuilt domain tools and auxiliary pretrained perception or grasp models are unavailable.
    \item \textbf{L1: perception and planning tools.} Explicit grants add installed tools for visual reconstruction, geometry, inverse kinematics, motion planning, symbolic solving, and observation-derived proxy dynamics or optimization. Geometry and dynamics estimates are derived from permitted observations and declared task parameters.
    \item \textbf{L2: grasp and contact models.} Explicit grants additionally permit pretrained models that generate or refine grasp and contact proposals, which the agent incorporates into its planning and control.
    \item \textbf{L3: privileged observations.} A declared read-only profile additionally exposes selected simulator fields, such as object poses, velocities, geometry, or contacts, according to the environment's observation contract.
\end{itemize}

Each run records the observation profile, capability level, exact tool grants, and available tool versions. RGB-D observations require an authorized profile and a supporting environment. A higher level makes tools eligible for use; the actual tool set is fixed explicitly. Privileged read access preserves the same action interface and task rules. The evaluated model retains its declared frozen weights; additional pretrained action controllers and reference trajectories are excluded. Agents receive public task parameters, including launcher spring stiffness, while private initialization records and undeclared evaluator diagnostics remain with the evaluation host.

\subsection{Trial Design}
\label{sec:trials}
An evaluation group fixes the task, base seed, initial mode, operation and observation profiles, capability level, tool grants, budgets, and protocol and implementation versions. The agent version, code, and instructions are also fixed before execution. Every group contains exactly four independently initialized episodes. The standard schedule uses base seed zero; additional base seeds define separate groups.

Launcher tasks have one evaluated group: \emph{loaded}, with the first projectile preloaded. Subsequent reloading remains part of any task that requires further shots. Each other task has one default group. Thus, evaluation of the full playground contains 410 groups and 1,640 trial conditions per fixed setting. EP-100 contains 100 groups and 400 conditions. EP-Easy and EP-Hard each contain 50 groups and 200 conditions, and EP-20 contains 20 groups and 80 conditions. These counts describe initial conditions, not the number of attempts under a Pass@2 budget. The runtime also supports ground initialization, but that mode is excluded from these benchmark scores.

The four conditions preserve each task's goal and rules while varying family-specific factors. Launcher trials use spring stiffnesses of 22, 24, 26, and 28\,N/m, together with prescribed launcher and ammunition offsets and visual backgrounds. Hanoi and pocket-cube trials use consecutive reset seeds with bounded pose, mass, friction, and lighting perturbations; bookshelf trials use seeded native layout sampling. Rubik's cube, Kongming lock, and ManiSkill tasks retain their authored physical configurations and use lighting multipliers of 0.85, 0.95, 1.05, and 1.15. Formal ManiSkill trials use physics variant zero. Exact conditions are specified in the per-task schedule.

Each trial starts with a fresh environment and agent session. Within an episode, the agent may retain observations and adapt its plan, while physical changes, damage, and consumed resources persist. Between episodes, agent memory, workspaces, and caches are cleared; trajectories and private outcomes are not reused for adaptation. Initialization and reset belong to the evaluation host. All four registered conditions remain in the record. In the base single-attempt protocol, completed failures or invalid attempts cannot be replaced by additional retries. A declared multi-attempt experiment must retain each attempt and specify its opportunity budget and infrastructure-exclusion policy separately; Section~\ref{sec:experiment-pass} defines the Codex Pass@1/Pass@2 extension.

\subsection{Resource Budgets}
\label{sec:budgets}
Table~\ref{tab:episode-budgets} gives the registered execution limits. Actions advance simulation at the native control rate, with physics evaluated at the corresponding substep frequency. Observation and tool reads leave simulated time unchanged. A separate supervisor enforces elapsed wall time, including startup and initialization. Dynamics tasks can additionally impose event deadlines of 5--20\,s; the required successful dwell must finish within the applicable deadline.

\begin{table}[htbp]
\centering
\caption{Per-episode execution budgets. Times are in seconds. The bookshelf limit depends on the task configuration. Task-specific event deadlines apply in addition to these episode limits.}
\label{tab:episode-budgets}
\small
\begin{tabularx}{\linewidth}{@{}Y r r r r@{}}
\toprule
\eptablehead
\textbf{Task family} & \shortstack{\textbf{Physics/control}\\\textbf{(Hz)}} & \shortstack{\textbf{Maximum}\\\textbf{actions}} & \shortstack{\textbf{Simulated}\\\textbf{time}} & \shortstack{\textbf{Wall}\\\textbf{time}} \\
\midrule
Launcher & 1,000/50 & 7,500 & 150 & 600 \\
Hanoi & 120/30 & 18,000 & 600 & 3,600 \\
Pocket cube & 240/60 & 18,000 & 300 & 3,600 \\
Rubik's cube & 120/30 & 109,800 & 3,660 & 19,800 \\
Kongming lock & 120/30 & 36,000 & 1,200 & 3,600 \\
Bookshelf & 120/30 & 1,500--12,300 & 50--410 & 3,600 \\
ManiSkill: long horizon & 240/20 & 6,000 & 300 & 3,600 \\
ManiSkill: geometry/dynamics & 240/20 & 2,400 & 120 & 3,600 \\
\bottomrule
\end{tabularx}
\end{table}

Model-side budgets, including token consumption, image inputs, and auxiliary computation, are specified by the experiment host separately from the environment limits. Comparisons use the same declared allowances and report realized usage. The environment gateway meters physical execution; model-resource accounting is performed by the agent evaluation system.

\subsection{Episode Scoring}
\label{sec:scoring}
Success is determined by the task's independent evaluator from the executed physical trajectory. The final state must satisfy the goal, the interaction history must satisfy task rules, and all required physical, release, and stability conditions must hold before the applicable budget expires. Recorded violations remain binding throughout the episode. Calling \mbox{\texttt{finish}} evaluates existing evidence and grants no additional simulation or settling time.

The success predicates reflect each task's structure. Launcher tasks require the specified targets, impact order, protection rules, and any required fracture, toppling, or damage paths; protected-target scenes include a minimum 3\,s observation period after the last required-target hit. Hanoi requires the exact ring arrangement, legal moves, valid seating, and 1\,s stability. Pocket cubes require the target state and prescribed number of committed quarter turns, aligned detents, no active layer, and 1\,s stability. Rubik's cubes require a solved physical configuration, legal fixture use, return and release, and a fresh 0.5\,s stable window. Kongming locks require the case-specific arrangement and release, followed by 2\,s stability for assembly or 1\,s for disassembly. Bookshelf tasks require all placement rules to hold continuously for 0.5\,s. ManiSkill tasks require the ordered milestones, final predicates, protected-object conditions, and 1\,s successful dwell. Numerical tolerances remain fixed in each task contract.

Each registered trial is classified as \emph{success}, \emph{failure}, \emph{invalid}, or \emph{missing}. A valid failure includes an unsuccessful simulated-time timeout, object loss, physical-rule violation, or voluntary termination without success. Initialization or evaluator errors, rejected actions, agent exceptions, and wall-clock expiration produce invalid attempts. A slot without a terminal record is missing. These categories preserve the distinction between an unsuccessful physical attempt and an incomplete evaluation record.

\subsection{Evaluation Metrics}
\label{sec:metrics}
A group is complete when all four slots contain valid terminal outcomes. For a complete group $g$, let $z_{g,i}\in\{0,1\}$ denote success in trial $i$. We report its success rate and all-four-success indicator:
\begin{equation}
    S_g=\frac{1}{4}\sum_{i=1}^{4}z_{g,i},
    \qquad
    A_g=\prod_{i=1}^{4}z_{g,i}.
    \label{eq:group-metrics}
\end{equation}
The fixed denominator includes every successful or failed trial. If any slot is invalid or missing, the group is marked partial and both final metrics are undefined. Its observed success count $n_g^{+}$, invalid and missing trial indices, and lower bound $n_g^{+}/4$ are retained explicitly.

Group metrics are combined in three stages so that each suite carries equal weight. A task $\tau$ with registered groups $\mathcal{G}_\tau$ contributes to its suite $k$ and to the evaluated subset through
\begin{equation}
    S_\tau=\frac{1}{|\mathcal{G}_\tau|}\sum_{g\in\mathcal{G}_\tau}S_g,
    \qquad
    S_k=\frac{1}{|\mathcal{T}_k|}\sum_{\tau\in\mathcal{T}_k}S_\tau,
    \qquad
    S=\frac{1}{5}\sum_{k=1}^{5}S_k,
    \label{eq:aggregate-metrics}
\end{equation}
where $\mathcal{T}_k$ is the set of subset tasks in suite $k$. At a fixed evaluation setting, each task has one group, so its task score equals $S_g$, including for loaded-mode launcher tasks. The suite score averages the tasks of one suite, and the subset score assigns one fifth of its weight to each of the five suites, so the reported score of EP-100, EP-Easy, EP-Hard, or EP-20 reflects every suite equally regardless of its size. The all-four indicator is aggregated in the same way and yields $A_\tau$, $A_k$, and $A$. We also organize results by task type and embodiment, keeping different access settings and attempt budgets distinct. Partial evaluations retain the planned denominator, observed-success lower bound, completed-group coverage, and invalid or missing counts, with final aggregate metrics left undefined.

Success rates are accompanied by efficiency and failure statistics. Time@S is the mean wall-clock time from episode start to the success event over successful trials, and Tok@S is the mean number of model tokens that a successful trial consumes, counting cached and uncached input together with output. Each failed trial receives one failure category from its first recorded violation or from its termination reason, and the share of each category describes how agents fail. Table~\ref{tab:failure-codes} in Appendix~\ref{app:ep100} lists the categories, including the separate category for invalid attempts.

Each episode record includes the task and group identity, trial index, agent and environment revisions, observation and tool settings, budgets, realized initialization, action count, elapsed simulated and wall time, terminal outcome, and first recorded violation. Detailed state and scorer logs are retained by the evaluator; agents receive only authorized observations and filtered results. These records support reproducible comparisons and trajectory-level analysis of perception, planning, and execution.

\section{Experimental Evaluation}
\label{sec:experiments}

We evaluate GPT-6-Astra and Opus-5.5-extra on EP-100.

\subsection{Experimental Setup}
\paragraph{Evaluation configuration}
We evaluate Codex using \texttt{gpt-6-astra} with high reasoning effort and L3 privileged observations. Opus-5.5-extra uses extra reasoning effort. The agents generate and execute control programs through the public environment SDK, using Python's standard library, NumPy, and Pillow. No auxiliary tool grants, reference solutions, or pretrained action controllers are provided. Each attempt follows the physical execution budgets in Table~\ref{tab:episode-budgets}, with a declared allowance of 30 million tokens, including cached input, and 100 image inputs.

The evaluation covers all 100 EP-100 tasks under four prescribed conditions at base seed zero, yielding 400 task--condition pairs. Launcher tasks use loaded initialization only: the first projectile is preloaded, while subsequent loading remains part of multi-shot tasks. EP-Easy, EP-Hard, and EP-20 are evaluated by selecting the corresponding records under the loaded-only definitions in Section~\ref{sec:ep100-difficulty}.

\paragraph{Attempt eligibility}
Each condition is scored using its first two valid attempts in chronological order, stopping at the first success. For GPT-6-Astra, the final audit retains 846 historical records: 736 scored attempts, comprising 91 successes and 645 failures; 69 confirmed infrastructure incidents excluded from the opportunity budget; and 41 extra attempts outside the scoring window. Those extra attempts include two native successes, which remain unscored. Diagnostic replays are used only for auditing. All 400 conditions have complete outcomes, with no outstanding valid opportunities. Ordinary agent failures and budget exhaustion, including wall-clock timeouts, consume an opportunity; confirmed infrastructure incidents do not. These rules extend the base single-attempt protocol.

Recorded process launches establish separate agent threads and public, scratch, and runtime directories for all 736 scored attempts. Trial identity is supported by 548 standalone manifests and 188 delegated native trace manifests. This supports isolation at the session and directory level, but does not certify cleared host or service caches or complete simulator-state independence. Native verdicts are retained, including failures involving incompletely disclosed rules, with disclosure defects reported separately below.

\paragraph{Metrics}
\label{sec:experiment-pass}
Pass@1 measures success on the first valid attempt; Pass@2 measures success within the first two valid attempts on the same condition. A first-attempt success contributes to both metrics. Let $p^{(k)}_{\tau,i}$ indicate success within $k$ valid attempts for task $\tau$ and condition $i$. For subset $\mathcal{D}$, we compute
\begin{equation}
    \operatorname{Pass@}k(\mathcal{D})=
    \frac{1}{5}\sum_{s=1}^{5}
    \frac{1}{|\mathcal{D}_s|}\sum_{\tau\in\mathcal{D}_s}
    \frac{1}{4}\sum_{i=1}^{4}p^{(k)}_{\tau,i},
    \qquad k\in\{1,2\},
    \label{eq:experiment-pass}
\end{equation}
where $\mathcal{D}_s$ contains the tasks in suite $s$. Conditions are averaged within tasks, tasks within suites, and the five suites equally. The four conditions are separate evaluation instances, not additional attempts at one instance. Pass@2 is computed from observed outcomes rather than inferred from Pass@1.

\subsection{Main Results}
\paragraph{Performance across attempts}
GPT-6-Astra achieves 13.80\% Pass@1 and 19.94\% Pass@2 on EP-100 (Table~\ref{tab:model-subsets}), a gain of 6.15 percentage points before rounding. Opus-5.5-extra achieves 29.22\% Pass@1 and 35.85\% Pass@2 using 683 valid attempts. For GPT-6-Astra, the first valid attempt solves 64 of 400 conditions; second attempts add 27 successes, recovering 8.04\% of the 336 initially unsuccessful conditions. The pooled condition rates are 16.00\% and 22.75\%, respectively, whereas the headline scores weight each suite equally. Thus, a second opportunity improves coverage, but 309 GPT-6-Astra conditions remain unsuccessful.

\begin{table}[htbp]
\centering
\caption{GPT-6-Astra and Opus-5.5-extra results on the four EP subsets. Pass@1 and Pass@2 are suite-balanced percentages. Attempts count valid opportunities used. EP-Easy and EP-Hard partition EP-100, while EP-20 overlaps both and uses the same records. Launch is loaded-only throughout.}
\label{tab:model-subsets}
\label{tab:codex-subsets}
\small
\setlength{\tabcolsep}{4pt}
\begin{tabularx}{\linewidth}{@{}l Y r r r r r@{}}
\toprule
\eptablehead
\textbf{Subset} & \textbf{Model} & \textbf{Tasks} & \textbf{Conditions} & \textbf{Attempts} & \textbf{Pass@1} & \textbf{Pass@2} \\
\midrule
EP-100 & GPT-6-Astra & 100 & 400 & 736 & 13.80 & 19.94 \\
 & Opus-5.5-extra & 100 & 400 & 683 & 29.22 & 35.85 \\
EP-Easy & GPT-6-Astra & 50 & 200 & 358 & 18.42 & 27.41 \\
 & Opus-5.5-extra & 50 & 200 & 332 & 33.99 & 42.88 \\
EP-Hard & GPT-6-Astra & 50 & 200 & 378 & 8.81 & 12.01 \\
 & Opus-5.5-extra & 50 & 200 & 351 & 24.21 & 28.48 \\
EP-20 & GPT-6-Astra & 20 & 80 & 147 & 15.42 & 18.75 \\
 & Opus-5.5-extra & 20 & 80 & 137 & 30.58 & 39.13 \\
\bottomrule
\end{tabularx}
\end{table}

\paragraph{Subset performance}
For GPT-6-Astra, EP-Easy reaches 18.42\% Pass@1 and 27.41\% Pass@2, compared with 8.81\% and 12.01\% on EP-Hard. The second-attempt gains are 8.99 and 3.20 percentage points, respectively. The workload-based split therefore separates aggregate performance in this evaluation, although the relationship is not uniform: Home scores higher on EP-Hard than on EP-Easy (Table~\ref{tab:model-suites}). EP-20 obtains 15.42\% Pass@1 and 18.75\% Pass@2 over 80 conditions. Its suite profile differs from EP-100, especially for Tower and Home. As a curated selection partly informed by pilot success evidence, EP-20 provides a compact diagnostic evaluation rather than an unbiased estimate of full-set performance.

\begin{table}[htbp]
\centering
\caption{Suite-level success rates (\%). Each cell gives Pass@1 / Pass@2. Conditions are averaged within tasks and tasks within suites. The subset scores in Table~\ref{tab:model-subsets} average the five suite entries.}
\label{tab:model-suites}
\label{tab:codex-suites}
\small
\setlength{\tabcolsep}{4pt}
\begin{tabularx}{\linewidth}{@{}l Y r r r r@{}}
\toprule
\eptablehead
\textbf{Suite} & \textbf{Model} & \textbf{EP-100} & \textbf{EP-Easy} & \textbf{EP-Hard} & \textbf{EP-20} \\
\midrule
Launch & GPT-6-Astra & 19.12 / 25.74 & 20.59 / 30.88 & 17.65 / 20.59 & 12.50 / 16.67 \\
 & Opus-5.5-extra & 15.51 / 27.70 & 15.21 / 32.60 & 15.81 / 22.81 & 5.65 / 19.78 \\
Tower & GPT-6-Astra & 29.76 / 41.67 & 45.45 / 61.36 & 12.50 / 20.00 & 12.50 / 25.00 \\
 & Opus-5.5-extra & 60.59 / 66.14 & 71.51 / 79.82 & 48.58 / 51.09 & 50.00 / 56.25 \\
Lock & GPT-6-Astra & 6.25 / 14.06 & 12.50 / 28.13 & 0.00 / 0.00 & 0.00 / 0.00 \\
 & Opus-5.5-extra & 39.81 / 45.63 & 52.03 / 60.56 & 27.60 / 30.70 & 37.63 / 37.63 \\
Cube & GPT-6-Astra & 2.08 / 2.08 & 4.17 / 4.17 & 0.00 / 0.00 & 8.33 / 8.33 \\
 & Opus-5.5-extra & 6.04 / 10.72 & 8.18 / 15.03 & 3.90 / 6.41 & 14.82 / 27.41 \\
Home & GPT-6-Astra & 11.76 / 16.18 & 9.38 / 12.50 & 13.89 / 19.44 & 43.75 / 43.75 \\
 & Opus-5.5-extra & 24.17 / 29.06 & 23.03 / 26.41 & 25.17 / 31.41 & 44.80 / 54.58 \\
\bottomrule
\end{tabularx}
\end{table}

\paragraph{Performance across tasks}
For GPT-6-Astra, Tower has the highest suite-level Pass@2 at 41.67\%. Its Physics and Plan types reach 68.75\% and 50.00\%, while Dex reaches 9.38\%. Launch reaches 25.74\%, with Aim at 46.43\% and Rules at 5.56\%. Lock reaches 14.06\%, with all nine successful conditions in EP-Easy and no success in Lock-Dex. Cube records one successful Pocket condition and no Rubik successes, yielding 2.08\% overall. Home reaches 16.18\%, with successes confined to Precise and Dynamic; Tidy and Long record none. These differences describe the evaluated task configurations and do not constitute controlled comparisons of reasoning or embodiment.

For GPT-6-Astra, at least one condition succeeds for 31 tasks on the first valid attempt and 40 tasks within two attempts. Only four and nine tasks, respectively, succeed across all four conditions. Success therefore remains uneven across the prescribed variations, even with a second opportunity. The latter count permits two attempts per condition and is distinct from the base all-four single-attempt metric. Table~\ref{tab:model-outcomes} reports condition outcomes for both models and Opus-5.5-extra successful-attempt timing.

\begin{table}[htbp]
\centering
\caption{EP-100 condition outcomes and Opus-5.5-extra successful-attempt timing. Success and failure count conditions after at most two valid attempts, rather than individual attempts. Time@S is the mean wall-clock time of successful attempts in minutes and excludes prior unsuccessful attempts.}
\label{tab:model-outcomes}
\small
\setlength{\tabcolsep}{4pt}
\begin{tabularx}{\linewidth}{@{}Y r r r r r r@{}}
\toprule
\eptablehead
 & & \multicolumn{2}{c}{\textbf{GPT-6-Astra}} & \multicolumn{3}{c}{\textbf{Opus-5.5-extra}} \\
\cmidrule(lr){3-4}\cmidrule(lr){5-7}
\eptablehead
\textbf{Suite} & \textbf{Cond.} & \textbf{Success} & \textbf{Failure} & \textbf{Success} & \textbf{Failure} & \textbf{Time@S (min)} \\
\midrule
Launch & 136 & 35 & 101 & 36 & 100 & 6.2 \\
Tower & 84 & 35 & 49 & 55 & 29 & 38.5 \\
Lock & 64 & 9 & 55 & 20 & 44 & 37.0 \\
Cube & 48 & 1 & 47 & 5 & 43 & 32.5 \\
Home & 68 & 11 & 57 & 17 & 51 & 26.6 \\
\bottomrule
\end{tabularx}
\end{table}

\subsection{Evaluation Diagnostics}
\paragraph{Observed failure events}
The audit resolves the 399 previously unclassified failure records and assigns diagnostic labels to all 645 currently scored failures. Table~\ref{tab:codex-failures} summarizes the most frequent labels by suite. Launch failures frequently involve limit and interlock checks, including 86 elevation-limit events and 50 launches before completion of an aiming cycle. Tower failures include penetration, ring displacement, and native target-validation aborts. Lock is dominated by contact checks, while 64 of 94 Cube failures carry mechanism-constraint labels. Home failures include time limits, penetration, and object loss. These labels identify native termination events; they do not, by themselves, establish whether the agent, controller, or simulator caused the underlying physical behavior.

\begin{table}[htbp]
\centering
\caption{Frequent diagnostic labels among GPT-6-Astra scored failed attempts. Counts are incidences within each suite, not task success rates. Multiple labels may apply to one failure; only the most frequent labels are shown.}
\label{tab:codex-failures}
\small
\begin{tabularx}{\linewidth}{@{}l r Y@{}}
\toprule
\eptablehead
\textbf{Suite} & \textbf{Failed attempts} & \textbf{Frequent labels (count)} \\
\midrule
Launch & 211 & Limit (129), rule (58) \\
Tower & 108 & Penetration (35), displacement/loss (33), limit (27) \\
Lock & 115 & Contact (97), penetration (19) \\
Cube & 94 & Mechanism (64), limit (16) \\
Home & 117 & Time (40), penetration (27), displacement/loss (23) \\
\bottomrule
\end{tabularx}
\end{table}

\paragraph{Failure attribution}
The audit identifies incomplete rule or threshold disclosure affecting 574 of the 645 scored failures (88.99\%). Examples include launcher joint-speed margins, forbidden-contact thresholds in Lock, and fixed-joint integrity tolerances in Cube. A missing numerical tolerance is distinguished from an undisclosed task obligation. These verdicts remain in the reported scores under the audit policy; disclosure defects are not treated as successes or automatically excluded. Consequently, the scores measure completion under the implemented native contracts, while the failure distribution cannot be interpreted solely as a measure of agent reasoning or control deficiencies. Evaluations with fully disclosed contracts are needed to isolate those limitations.

\paragraph{Measurement scope}
Mean supervisor-recorded time over the 91 GPT-6-Astra successful attempts is approximately 6.25 minutes, measured at roughly 0.25-second polling resolution; it excludes prior unsuccessful attempts. Complete provider token usage is unavailable for this run, so Tok@S is not reported. Controlled access-level, controller, and feedback ablations are needed to separate agent capabilities from evaluation artifacts.

\section{Conclusion and Future Work}
\label{sec:conclusion}
We introduced Embodied Playground, a benchmark comprising 410 robotic task environments for evaluating spatial, physical, and strategic reasoning through embodied interaction. EP-100 and its complementary subsets support focused evaluation under a shared protocol with independent physical success checks. The audited Codex results show limited and uneven task completion under the implemented native contracts. The audit also identifies rule-disclosure gaps that constrain attribution of failures to agent capabilities.

We will maintain and expand the benchmark with new tasks and improved evaluation mechanisms, including clearer rule disclosure and stronger validation, to strengthen coverage, reliability, and reproducibility. We will also evaluate a broader range of foundation models and agent systems to characterize their capabilities and track progress in embodied problem solving.

\bibliographystyle{unsrtnat}
\bibliography{sample}

\clearpage
\appendix
\section{Task Catalog and Evaluation Subset Indices}
\label{app:tasks}

\subsection{Catalog Organization}
The full Embodied Playground contains 410 registered task environments. Each identifier has the form \texttt{family/embodiment/case}; an environment is counted once for each registered task--embodiment combination. Repeated trials and launcher initialization modes belong to the evaluation protocol and do not increase the catalog size. The catalog below follows the task manifest at Agent-bench revision \texttt{a58a288}. Compact ranges enumerate all cases, while variant tables describe their objectives and principal sources of difficulty.

The launcher, Hanoi, lock, cube, and household environments contribute 100, 120, 60, 80, and 50 tasks, respectively. Household environments comprise 20 bookshelf tasks and 30 ManiSkill tasks. Native difficulty labels describe the design of an individual task family; evaluation subset assignments are recorded separately.

\subsection{Launcher: Aiming and Physical Causality}
\label{app:launcher}
The launcher catalog contains \texttt{marble/dual-panda/A001--A100}, implemented in MuJoCo. Two Panda arms operate the launcher through physical controls: aiming handwheels, a brake, a spring mechanism, and ammunition loading. Finite ammunition couples trajectory selection with decisions about projectile type, draw, and firing order. The runtime supports loaded and ground initialization of the same authored scene with different loading requirements. All EP benchmark subsets use loaded initialization only, with the first projectile preloaded; subsequent reloading remains part of multi-shot tasks.

The scenes range from unobstructed targets to interactions mediated by rolling objects, dominoes, bridges, roofs, and collapsing structures. Glass introduces fracture thresholds; protected targets and required impact sequences constrain otherwise successful shots. For example, A007 introduces draw control with a nearby target, A002 restricts the firing corridor with a bunker, A036 requires clearing an upper target while preserving one below, and A099 requires contact with an inclined return bank before either target can be cleared. These tasks combine geometric aiming with prediction of delayed physical consequences and recovery within a finite shot budget.

{\small
\begin{longtable}{@{}>{\raggedright\arraybackslash}p{0.07\linewidth}>{\raggedright\arraybackslash}p{0.08\linewidth}>{\raggedright\arraybackslash}p{0.79\linewidth}@{}}
\caption{Complete launcher index by native tier. Case numbers are not ordered by tier.}\\
\toprule
\textbf{Tier} & \textbf{Tasks} & \textbf{Case IDs} \\
\midrule\endfirsthead
\toprule
\textbf{Tier} & \textbf{Tasks} & \textbf{Case IDs} \\
\midrule\endhead
\bottomrule\endfoot
1 & 4 & A007, A008, A009, A010 \\
2 & 9 & A001, A011, A012, A013, A017, A018, A023, A037, A075 \\
3 & 25 & A002, A003, A004, A014, A019, A020, A024, A025, A026, A038, A039, A040, A044, A049, A050, A053, A054, A056, A059, A061, A064, A071, A076, A077, A080 \\
4 & 40 & A005, A006, A015, A016, A021, A022, A027, A028, A029, A030, A041, A042, A043, A045, A048, A051, A052, A055, A057, A058, A060, A062, A063, A065, A066, A069, A070, A072, A073, A078, A079, A081, A082, A083, A084, A085, A086, A090, A091, A094 \\
5 & 22 & A031, A032, A033, A034, A035, A036, A046, A047, A067, A068, A074, A087, A088, A089, A092, A093, A095, A096, A097, A098, A099, A100 \\
\end{longtable}
}

\subsection{Hanoi: Planning with Geometric and Physical Constraints}
\label{app:hanoi}
Forty board configurations are instantiated with three embodiments: \nolinkurl{hanoi_panda/panda}, \nolinkurl{hanoi_sharpa/fr3-sharpa}, and \nolinkurl{hanoi_floating/floating-sharpa}. Each prefix contains cases H01--H40, yielding 120 Isaac Lab environments. The board contains five rings and three pegs. An agent must reach the specified arrangement by moving only an exposed ring and preserving the size ordering on every peg, while grasping, transporting, and seating rings physically.

The configurations form ten groups of four cases. Within each group, the cases have native levels 1--4 and shortest legal solution lengths of 1, 3, 9, and 31 moves, respectively. The groups vary spatial layout, contact conditions, or motion constraints. This structure provides related planning problems across parallel grippers and dexterous hands, while exposing the interaction between a legal symbolic plan and its physical execution.

{\small
\begin{longtable}{@{}>{\raggedright\arraybackslash}p{0.15\linewidth}>{\raggedright\arraybackslash}p{0.25\linewidth}>{\raggedright\arraybackslash}p{0.54\linewidth}@{}}
\caption{Hanoi configuration index. Each row contributes four cases per embodiment and twelve environments in total.}\\
\toprule
\textbf{Cases} & \textbf{Variation} & \textbf{Principal demand} \\
\midrule\endfirsthead
\toprule
\textbf{Cases} & \textbf{Variation} & \textbf{Principal demand} \\
\midrule\endhead
\bottomrule\endfoot
H01--H04 & State decomposition & Identify the required subproblem and maintain legal ring ordering. \\
H05--H08 & Planar layout & Adapt transfers to different peg arrangements. \\
H09--H12 & Support height & Transfer rings between supports at different elevations. \\
H13--H16 & Insertion clearance & Align ring holes and pegs under varying clearance. \\
H17--H20 & Mass distribution & Adapt lifting and transport to ring masses. \\
H21--H24 & Friction & Maintain stable grasps under different contact conditions. \\
H25--H28 & Handle orientation & Approach and grasp handles with varying yaw. \\
H29--H32 & Path obstacles & Route ring transfers around intervening obstacles. \\
H33--H36 & Peg length & Clear and seat rings on pegs of different lengths. \\
H37--H40 & Combined constraints & Coordinate planning with multiple geometric and physical variations. \\
\end{longtable}
}

\subsection{Cubes: Layer Manipulation and Sequential Solving}
\label{app:cubes}
\paragraph{Pocket cube.}
Ten prescribed transitions on a $2\times2$ cube are instantiated under \texttt{pocket\_cube/aloha} and \texttt{pocket\_cube/dual-sharpa}, giving 20 Isaac Lab environments. The latter uses two floating Sharpa hands. Each case specifies an initial cube state and a target transition. A face-order interlock constrains the sequence, and success requires the target state, the prescribed number of committed quarter turns, and aligned, stable layers. These tasks emphasize support, torque application, regrasping, and coordination between the manipulating and stabilizing hands. Table~\ref{tab:pocket-index} lists all ten transitions; the native level is shared across the two embodiments.

{\small
\begin{longtable}{@{}>{\raggedright\arraybackslash}p{0.1\linewidth}>{\raggedright\arraybackslash}p{0.07\linewidth}>{\raggedright\arraybackslash}p{0.23\linewidth}>{\raggedright\arraybackslash}p{0.52\linewidth}@{}}
\caption{Pocket-cube case index. Standard face notation is used; a prime denotes an inverse turn and 2 denotes a half turn.\label{tab:pocket-index}}\\
\toprule
\textbf{Case} & \textbf{Level} & \textbf{Transition} & \textbf{Manipulation demand} \\
\midrule\endfirsthead
\toprule
\textbf{Case} & \textbf{Level} & \textbf{Transition} & \textbf{Manipulation demand} \\
\midrule\endhead
\bottomrule\endfoot
C01 & 1 & \texttt{U} & Support the lower layer during a top-layer turn. \\
C03 & 1 & \texttt{L} & Create clearance for a left-layer turn. \\
C07 & 2 & \texttt{D} & Free the bottom layer from its support. \\
C08 & 2 & \texttt{U2} & Complete a half turn through intermediate detents. \\
C09 & 3 & \texttt{L} & Control a side-layer turn with increased mass. \\
C10 & 3 & \texttt{R'} & Maintain contact during a low-friction reverse turn. \\
C14 & 4 & \texttt{U F R} & Coordinate successive turns about three axes. \\
C15 & 4 & \texttt{F' U' R'} & Execute a reverse sequence to restore the target state. \\
C17 & 5 & \texttt{R U R' U'} & Maintain state tracking through a commutator sequence. \\
C20 & 5 & \texttt{B' U' R' F'} & Recover through a multi-axis sequence with combined perturbations. \\
\end{longtable}
}

\paragraph{Rubik's cube.}
Thirty scrambled $3\times3$ configurations are instantiated under \texttt{rubiks\_panda/panda} and \texttt{rubiks\_sharpa/floating-sharpa}, giving 60 Isaac Lab environments. Each case fixes a scramble, pose, and logical orientation. Scramble length is measured in construction tokens: a half turn counts as one token but two quarter turns. The token count therefore describes the construction procedure rather than the shortest solution length.

An agent must infer the cube state, plan a solving sequence, and execute layer turns through the declared grasp attachment and detent mechanics of \texttt{hybrid\_fixture\_v1}. Completion requires a solved cube, return to the designated home configuration, release, and stability. Pose and orientation variations require agents to maintain a consistent mapping between observed faces and planned moves.

{\small
\begin{longtable}{@{}>{\raggedright\arraybackslash}p{0.2\linewidth}>{\raggedright\arraybackslash}p{0.23\linewidth}>{\raggedright\arraybackslash}p{0.25\linewidth}>{\raggedright\arraybackslash}p{0.24\linewidth}@{}}
\caption{Rubik's cube index. Each case range appears under both embodiment prefixes.}\\
\toprule
\textbf{Cases} & \textbf{Native level} & \textbf{Scramble tokens} & \textbf{Environments} \\
\midrule\endfirsthead
\toprule
\textbf{Cases} & \textbf{Native level} & \textbf{Scramble tokens} & \textbf{Environments} \\
\midrule\endhead
\bottomrule\endfoot
C01--C10 & Easy & 5 & 20 \\
C11--C20 & Medium & 10 & 20 \\
C21--C30 & Hard & 15 & 20 \\
\end{longtable}
}

\subsection{Kongming Lock: Assembly and Disassembly}
\label{app:lock}
The lock catalog comprises a six-piece burr puzzle with five notched bars and one solid key. Thirty cases are instantiated under \texttt{kongming\_lock/dual-panda} and \texttt{kongming\_lock/dual-floating-sharpa}, giving 60 Isaac Lab environments. Case prefixes A and D denote assembly and disassembly, respectively. Each direction contains complete-lock, key-only, and key-plus-bar objectives, allowing the catalog to vary the extent of a manipulation sequence independently of the initial pose.

Each five-case block spans poses P01--P05 in order. Their native pose difficulty labels are Easy, Easy, Medium, Hard, and Hard. For partial tasks, the key-plus-bar objective concerns the key and \texttt{frame\_y1}; complete tasks concern all six pieces. Assembly requires physically consistent insertion and a stable released structure. Disassembly requires extracting the designated pieces and placing them in their prescribed final arrangement. Both directions test ordering dependencies, spatial clearance, and coordinated support during contact-rich motion.

{\small
\begin{longtable}{@{}>{\raggedright\arraybackslash}p{0.18\linewidth}>{\raggedright\arraybackslash}p{0.2\linewidth}>{\raggedright\arraybackslash}p{0.34\linewidth}>{\raggedright\arraybackslash}p{0.2\linewidth}@{}}
\caption{Complete lock index. Every listed range is instantiated with both embodiments.}\\
\toprule
\textbf{Assembly} & \textbf{Disassembly} & \textbf{Objective} & \textbf{Environments} \\
\midrule\endfirsthead
\toprule
\textbf{Assembly} & \textbf{Disassembly} & \textbf{Objective} & \textbf{Environments} \\
\midrule\endhead
\bottomrule\endfoot
A01--A05 & D01--D05 & Complete six-piece lock & 20 \\
A06--A10 & D06--D10 & Key only & 20 \\
A11--A15 & D11--D15 & Key and one bar & 20 \\
\end{longtable}
}

\subsection{Bookshelf: Organization and Scene Correction}
\label{app:bookshelf}
Twenty Isaac Lab tasks use \texttt{bookshelf/dual-panda}. Their case IDs combine one of five setting names with a native level D1--D4. The agent organizes a desk hutch according to object-specific placement rules: books stand upright and form a packed row on the middle shelf; display objects stand on the top shelf with the required spacing and orientation; crates or boxes occupy the bottom region; and trash belongs in the waste bin. Completion requires released, stationary objects in valid placements.

The settings vary the starting container, object composition, and need to correct an existing arrangement. Higher native levels introduce additional objects, flat or awkward initial poses, named-wall packing, clutter, or misplaced objects. In correction tasks, the agent must identify which existing placements violate the rules from the scene. The task names below summarize all twenty cases.

{\small
\begin{longtable}{@{}>{\raggedright\arraybackslash}p{0.3\linewidth}>{\raggedright\arraybackslash}p{0.66\linewidth}@{}}
\caption{Bookshelf case index. All IDs have the prefix \texttt{bookshelf/dual-panda/}.}\\
\toprule
\textbf{Case} & \textbf{Task configuration} \\
\midrule\endfirsthead
\toprule
\textbf{Case} & \textbf{Task configuration} \\
\midrule\endhead
\bottomrule\endfoot
\texttt{BooksTableD1} & Three upright books onto the middle shelf \\
\texttt{BooksTableD2} & Five books, some lying flat \\
\texttt{BooksTableD3} & Books onto the middle shelf \\
\texttt{BooksTableD4} & Ten books against a named wall \\
\texttt{ShelfTableD1} & Books, a trophy and one box \\
\texttt{ShelfTableD2} & Books, three display objects and two boxes \\
\texttt{ShelfTableD3} & Organize a full desk hutch from the desk \\
\texttt{ShelfTableD4} & A crowded desk with flat books and awkward display objects \\
\texttt{ShelfCrateD1} & Unpack a neatly packed crate \\
\texttt{ShelfCrateD2} & Unpack a small jumbled crate \\
\texttt{ShelfCrateD3} & Unpack a jumbled moving crate onto the hutch \\
\texttt{ShelfCrateD4} & Unpack a full crate against a named wall \\
\texttt{ShelfCrateFixD1} & One book on the wrong layer of a tidy shelf \\
\texttt{ShelfCrateFixD2} & Unpack a packed crate and correct a misplaced globe \\
\texttt{ShelfCrateFixD3} & Unpack the crate and correct a misplaced globe \\
\texttt{ShelfCrateFixD4} & Unpack the crate and correct two misplaced objects \\
\texttt{TableCleanD1} & Three books and two pieces of trash \\
\texttt{TableCleanD2} & Books, display objects and four pieces of trash \\
\texttt{TableCleanD3} & Organize the hutch and throw the trash away \\
\texttt{TableCleanD4} & Organize the hutch with trash already hidden on it \\
\end{longtable}
}

\subsection{ManiSkill: Household Sequences, Precision, and Dynamics}
\label{app:maniskill}
Thirty environments run in ManiSkill/SAPIEN: \texttt{maniskill\_long\_horizon} contains L1--L10, \texttt{maniskill\_geometry} contains P1--P10, and \texttt{maniskill\_dynamics} contains D1--D10. Each case has one registered embodiment, listed below using its exact runtime identifier. The robot collection includes Fetch, Panda, and the three arm--hand combinations registered as \texttt{everyday\_robot\_01}, \texttt{everyday\_robot\_03}, and \texttt{everyday\_robot\_09}. Together, the family prefix, robot identifier, and case ID recover the complete environment identifier.

\paragraph{Long-horizon tasks.}
These tasks require object selection, ordered interaction with storage mechanisms, and restoration of the workspace. Goals specify both what to change and which objects to preserve. Dependencies include retracting a rack before closing a door, inserting dividers before packing, and clearing obstructions before closing a case. Ordered milestones and final-state predicates capture these intermediate and terminal requirements.

{\small
\begin{longtable}{@{}>{\raggedright\arraybackslash}p{0.07\linewidth}>{\raggedright\arraybackslash}p{0.28\linewidth}>{\raggedright\arraybackslash}p{0.59\linewidth}@{}}
\caption{Long-horizon household task index.}\\
\toprule
\textbf{Case} & \textbf{Robot identifier} & \textbf{Objective} \\
\midrule\endfirsthead
\toprule
\textbf{Case} & \textbf{Robot identifier} & \textbf{Objective} \\
\midrule\endhead
\bottomrule\endfoot
L1 & \texttt{fetch} & Clear a writing area, store books and stationery, close the drawer, and preserve a colleague's objects. \\
L2 & \texttt{everyday\_robot\_09} & Store ingredients and utensils, retract the oven rack, switch off both knobs, and close storage without disturbing the loaf. \\
L3 & \texttt{fetch} & Arrange two mug--spoon settings, clear the serving area, close storage, and preserve the third setting. \\
L4 & \texttt{everyday\_robot\_03} & Select the specified screwdriver, handle, and screws; pack, close, and latch the repair case. \\
L5 & \texttt{fetch} & Unload specified mugs and cutlery, preserve the bowl, retract the rack, and close the dishwasher and storage. \\
L6 & \texttt{fetch} & Pack bottles below a divider and food above it; preserve the separate lunch and close storage. \\
L7 & \texttt{fetch} & Select exactly three colored markers and a drawing book, store blocks, and preserve excluded items. \\
L8 & \texttt{fetch} & Select a suitable planter and saucer, arrange supplies, clear the work area, and preserve existing plants. \\
L9 & \texttt{everyday\_robot\_01} & Insert dividers, pack camera equipment by compartment, clear the lid rim, and latch the case. \\
L10 & \texttt{everyday\_robot\_09} & Move a box and mug to access a book, deliver the book, and restore the displaced objects. \\
\end{longtable}
}

\paragraph{Geometric precision.}
These tasks emphasize the orientation and clearance required for insertion, extraction, hooking, and rotation. Neighboring objects and constrained approach directions make the path to the goal part of the task. The washer task additionally requires acquiring a thin object from a flat surface and maintaining the requested orientation.

{\small
\begin{longtable}{@{}>{\raggedright\arraybackslash}p{0.07\linewidth}>{\raggedright\arraybackslash}p{0.28\linewidth}>{\raggedright\arraybackslash}p{0.59\linewidth}@{}}
\caption{Geometric precision task index.}\\
\toprule
\textbf{Case} & \textbf{Robot identifier} & \textbf{Objective} \\
\midrule\endfirsthead
\toprule
\textbf{Case} & \textbf{Robot identifier} & \textbf{Objective} \\
\midrule\endhead
\bottomrule\endfoot
P1 & \texttt{panda} & Seat round, square, and long pegs in their matching sockets without moving the board. \\
P2 & \texttt{everyday\_robot\_09} & Insert a key, turn it clockwise by a quarter turn, and withdraw it onto the mat. \\
P3 & \texttt{panda} & Align and seat a keyed connector while preserving neighboring connectors. \\
P4 & \texttt{everyday\_robot\_03} & Align an asymmetric lid with its corner tab and seat the entire rim. \\
P5 & \texttt{panda} & Guide a hook through a mug handle and release it in stable suspension. \\
P6 & \texttt{everyday\_robot\_09} & Extract a spatula through a cluttered holder without disturbing adjacent utensils. \\
P7 & \texttt{panda} & Insert an upright book into a narrow gap under a low shelf. \\
P8 & \texttt{everyday\_robot\_01} & Align, lift, and rotate a portafilter into its locked position. \\
P9 & \texttt{everyday\_robot\_09} & Align a dial with a narrow target tick while preserving adjacent controls. \\
P10 & \texttt{panda} & Acquire a thin washer, orient its face upright, and hold it steadily above the mat. \\
\end{longtable}
}

\clearpage
\paragraph{Dynamic interaction.}
These tasks require reasoning about motion, momentum, balance, or spring forces. Some impose a single intervention, an exact transferred quantity, or a completion deadline. The agent must anticipate the evolution of the scene after contact and use feedback to regulate motion while preserving protected objects or carried contents.

{\small
\begin{longtable}{@{}>{\raggedright\arraybackslash}p{0.07\linewidth}>{\raggedright\arraybackslash}p{0.28\linewidth}>{\raggedright\arraybackslash}p{0.59\linewidth}@{}}
\caption{Dynamic interaction task index.}\\
\toprule
\textbf{Case} & \textbf{Robot identifier} & \textbf{Objective} \\
\midrule\endfirsthead
\toprule
\textbf{Case} & \textbf{Robot identifier} & \textbf{Objective} \\
\midrule\endhead
\bottomrule\endfoot
D1 & \texttt{panda} & Stop a rolling orange on a safe mat before it leaves the counter. \\
D2 & \texttt{panda} & Position a catcher beneath a ball leaving a ramp and retain it after impact. \\
D3 & \texttt{panda} & Push a puck once so that it coasts to rest inside the scoring region. \\
D4 & \texttt{everyday\_robot\_01} & Transport a loaded tray while keeping cups upright and contents contained. \\
D5 & \texttt{panda} & Push a tall bottle around a guide block without tipping it. \\
D6 & \texttt{panda} & Close a loaded drawer within five seconds without disturbing its contents. \\
D7 & \texttt{everyday\_robot\_03} & Pour exactly eight of twelve beads and retain the remaining four in the source cup. \\
D8 & \texttt{panda} & Guide a spring-closing door shut without slamming or rebound. \\
D9 & \texttt{everyday\_robot\_01} & Tilt a maze to guide a ball through two gates while avoiding traps and direct contact. \\
D10 & \texttt{fetch} & Dock a wheeled cart accurately while keeping its load upright and contained. \\
\end{longtable}
}

\subsection{EP-100 Index}
\label{app:ep100}
All launcher entries use loaded initialization. Each EP-100 task carries a suite index such as Launch-01, a task type, and a split. Table~\ref{tab:ep-scores} defines the workload scores that rank tasks within each type, and Table~\ref{tab:failure-codes} lists the failure categories used in reporting. Table~\ref{tab:ep100-index} gives the complete index, in which a check mark denotes membership in EP-20.

\begin{table}[htbp]
\centering
\caption{Workload scores that rank tasks within each type. Higher scores are harder. Ties are broken by embodiment, with parallel grippers first, and then by task identifier.}
\label{tab:ep-scores}
\small
\begin{tabularx}{\linewidth}{@{}l Y@{}}
\toprule
\eptablehead
\textbf{Suite} & \textbf{Workload score} \\
\midrule
Launch & Ten times the native tier, plus the loaded-mode difficulty level D1--D5 recorded in the task atlas, plus the number of required targets \\
Tower & Twice the depth index, which is 0 to 3 for 1, 3, 9, and 31 moves, plus 1 for a physical theme, plus 2 for FR3--Sharpa or 3 for floating Sharpa \\
Lock & Subtask weight of 0 for the key, 2 for the key and bar, and 4 for the full lock, plus 1 for a hard pose, plus 0.5 for assembly, plus 2 for two floating Sharpa hands \\
Cube & Scramble or transition length in quarter turns, plus 3 for a Sharpa embodiment \\
Home & Number of objects for bookshelf tasks, and number of ordered milestones and final predicates for ManiSkill tasks \\
\bottomrule
\end{tabularx}
\end{table}

\begin{table}[htbp]
\centering
\caption{Failure categories. Each failed trial receives the category of its first recorded violation or of its termination reason.}
\label{tab:failure-codes}
\small
\begin{tabularx}{\linewidth}{@{}l Y@{}}
\toprule
\eptablehead
\textbf{Code} & \textbf{Meaning} \\
\midrule
\texttt{COLL} & Forbidden or excessive contact (robot-object, arm-arm, force limit) \\
\texttt{PEN} & Penetration beyond the audited limit \\
\texttt{LIMIT} & Joint, speed or physics limit exceeded \\
\texttt{DROP} & Object dropped, unseated or pushed out of the workspace \\
\texttt{RULE} & Task rule broken (illegal move, wrong order, protected item hit, launcher interlock) \\
\texttt{MECH} & Assist-mechanism constraint broken (cube fixture relation, mechanism command) \\
\texttt{AMMO} & All shots used without clearing the required targets \\
\texttt{TIME} & Environment step, simulated-time or wall-clock budget exhausted \\
\texttt{QUIT} & Agent finished without a qualified success \\
\texttt{BUDGET} & Agent token or image budget exhausted (orchestrator stop) \\
\texttt{ERR} & Invalid attempt caused by a runtime, evidence, or model-provider fault; it is recorded as invalid and leaves its group partial \\
\bottomrule
\end{tabularx}
\end{table}

{\footnotesize
\setlength{\tabcolsep}{3pt}
\begin{longtable}{@{}>{\raggedright\arraybackslash}p{0.1\linewidth}>{\raggedright\arraybackslash}p{0.11\linewidth}>{\raggedright\arraybackslash}p{0.30\linewidth}>{\raggedright\arraybackslash}p{0.07\linewidth}>{\raggedright\arraybackslash}p{0.05\linewidth}>{\raggedright\arraybackslash}p{0.3\linewidth}@{}}
\caption{Complete EP-100 index.}\label{tab:ep100-index}\\
\toprule
\textbf{Index} & \textbf{Type} & \textbf{Task} & \textbf{Split} & \textbf{EP-20} & \textbf{Environment identifier} \\
\midrule\endfirsthead
\toprule
\textbf{Index} & \textbf{Type} & \textbf{Task} & \textbf{Split} & \textbf{EP-20} & \textbf{Environment identifier} \\
\midrule\endhead
\bottomrule\endfoot
Launch-01 & Aim & Two-range ammunition planning & Hard &  & \nolinkurl{marble/dual-panda/A001} \\
Launch-02 & Aim & Single-target warm-up & Easy & \checkmark & \nolinkurl{marble/dual-panda/A007} \\
Launch-03 & Aim & Long-range energy budget & Easy &  & \nolinkurl{marble/dual-panda/A008} \\
Launch-04 & Aim & Elevated-platform shot & Easy &  & \nolinkurl{marble/dual-panda/A009} \\
Launch-05 & Aim & Toppling timber gate & Easy &  & \nolinkurl{marble/dual-panda/A010} \\
Launch-06 & Aim & Three-target fan & Hard &  & \nolinkurl{marble/dual-panda/A075} \\
Launch-07 & Aim & Left-low and right-high fork & Hard &  & \nolinkurl{marble/dual-panda/A031} \\
Launch-08 & Causal & Orchard rolling bodies & Easy & \checkmark & \nolinkurl{marble/dual-panda/A017} \\
Launch-09 & Causal & One beam, two targets & Easy &  & \nolinkurl{marble/dual-panda/A004} \\
Launch-10 & Causal & Light roof on three supports & Easy &  & \nolinkurl{marble/dual-panda/A039} \\
Launch-11 & Causal & Gravity-driven stone at a low fort & Easy &  & \nolinkurl{marble/dual-panda/A044} \\
Launch-12 & Causal & Whole arch and separate crown stone & Easy &  & \nolinkurl{marble/dual-panda/A071} \\
Launch-13 & Causal & Branching dominoes & Hard &  & \nolinkurl{marble/dual-panda/A006} \\
Launch-14 & Causal & Indirect clearance with two bridges & Hard &  & \nolinkurl{marble/dual-panda/A015} \\
Launch-15 & Causal & Breakable window in a fixed frame & Easy &  & \nolinkurl{marble/dual-panda/A029} \\
Launch-16 & Causal & Staggered twin towers & Hard &  & \nolinkurl{marble/dual-panda/A034} \\
Launch-17 & Causal & Two-ball relay & Hard &  & \nolinkurl{marble/dual-panda/A055} \\
Launch-18 & Causal & Upper-to-lower beam transfer & Hard &  & \nolinkurl{marble/dual-panda/A063} \\
Launch-19 & Causal & Movable flap at a narrow throat & Hard & \checkmark & \nolinkurl{marble/dual-panda/A072} \\
Launch-20 & Causal & Near target before collapse & Hard &  & \nolinkurl{marble/dual-panda/A073} \\
Launch-21 & Glass & Rigid glass village & Easy &  & \nolinkurl{marble/dual-panda/A018} \\
Launch-22 & Glass & Two glass layers & Easy & \checkmark & \nolinkurl{marble/dual-panda/A025} \\
Launch-23 & Glass & Low lintel, high target & Easy &  & \nolinkurl{marble/dual-panda/A064} \\
Launch-24 & Glass & One-shot glass breach & Hard &  & \nolinkurl{marble/dual-panda/A027} \\
Launch-25 & Glass & Glass-supported lower story & Hard &  & \nolinkurl{marble/dual-panda/A035} \\
Launch-26 & Rules & Far target before near target & Easy &  & \nolinkurl{marble/dual-panda/A026} \\
Launch-27 & Rules & Protect the garden neighbor & Easy &  & \nolinkurl{marble/dual-panda/A028} \\
Launch-28 & Rules & Fire again with damage retained & Easy &  & \nolinkurl{marble/dual-panda/A030} \\
Launch-29 & Rules & Attic target above a protected floor & Hard &  & \nolinkurl{marble/dual-panda/A036} \\
Launch-30 & Rules & Keep the lower chamber safe & Hard &  & \nolinkurl{marble/dual-panda/A047} \\
Launch-31 & Rules & Fragment-risk well & Hard & \checkmark & \nolinkurl{marble/dual-panda/A068} \\
Launch-32 & Rules & Numbered fan courtyard & Easy &  & \nolinkurl{marble/dual-panda/A091} \\
Launch-33 & Rules & Upper floor before lower floor & Hard &  & \nolinkurl{marble/dual-panda/A092} \\
Launch-34 & Rules & Material-ordered courtyard & Hard & \checkmark & \nolinkurl{marble/dual-panda/A093} \\
Tower-01 & Plan & Top-ring transfer & Easy &  & \nolinkurl{hanoi_panda/panda/H01} \\
Tower-02 & Plan & Local two-ring transfer & Easy &  & \nolinkurl{hanoi_panda/panda/H02} \\
Tower-03 & Plan & Reassemble a distributed state & Easy &  & \nolinkurl{hanoi_panda/panda/H03} \\
Tower-04 & Plan & Long five-ring rearrangement & Hard & \checkmark & \nolinkurl{hanoi_panda/panda/H04} \\
Tower-05 & Plan & Long combined challenge & Hard &  & \nolinkurl{hanoi_panda/panda/H40} \\
Tower-06 & Physics & Triangular transfer & Easy & \checkmark & \nolinkurl{hanoi_panda/panda/H06} \\
Tower-07 & Physics & Low-to-high platform & Easy &  & \nolinkurl{hanoi_panda/panda/H09} \\
Tower-08 & Physics & Medium-hole replacement & Hard &  & \nolinkurl{hanoi_panda/panda/H14} \\
Tower-09 & Physics & Light-ring lift & Easy &  & \nolinkurl{hanoi_panda/panda/H17} \\
Tower-10 & Physics & Standard-friction transfer & Hard &  & \nolinkurl{hanoi_panda/panda/H22} \\
Tower-11 & Physics & Single tilted handle & Easy &  & \nolinkurl{hanoi_panda/panda/H25} \\
Tower-12 & Physics & Long and short peg pair & Hard &  & \nolinkurl{hanoi_panda/panda/H34} \\
Tower-13 & Physics & Offset vertical barrier & Hard &  & \nolinkurl{hanoi_panda/panda/H31} \\
Tower-14 & Dex & Top-ring transfer & Easy &  & \nolinkurl{hanoi_floating/floating-sharpa/H01} \\
Tower-15 & Dex & Local two-ring transfer & Hard & \checkmark & \nolinkurl{hanoi_floating/floating-sharpa/H02} \\
Tower-16 & Dex & Low-to-high platform & Hard &  & \nolinkurl{hanoi_floating/floating-sharpa/H09} \\
Tower-17 & Dex & Single tilted handle & Hard &  & \nolinkurl{hanoi_floating/floating-sharpa/H25} \\
Tower-18 & Dex & Top-ring transfer & Easy & \checkmark & \nolinkurl{hanoi_sharpa/fr3-sharpa/H01} \\
Tower-19 & Dex & Local two-ring transfer & Hard &  & \nolinkurl{hanoi_sharpa/fr3-sharpa/H02} \\
Tower-20 & Dex & Low-to-high platform & Easy &  & \nolinkurl{hanoi_sharpa/fr3-sharpa/H09} \\
Tower-21 & Dex & Single tilted handle & Easy &  & \nolinkurl{hanoi_sharpa/fr3-sharpa/H25} \\
Lock-01 & Assemble & Assemble full lock & Hard &  & \nolinkurl{kongming_lock/dual-panda/A01} \\
Lock-02 & Assemble & Assemble key & Easy &  & \nolinkurl{kongming_lock/dual-panda/A06} \\
Lock-03 & Assemble & Assemble key + bar & Easy & \checkmark & \nolinkurl{kongming_lock/dual-panda/A11} \\
Lock-04 & Assemble & Assemble full lock & Hard &  & \nolinkurl{kongming_lock/dual-panda/A04} \\
Lock-05 & Assemble & Assemble key & Easy &  & \nolinkurl{kongming_lock/dual-panda/A09} \\
Lock-06 & Assemble & Assemble key + bar & Hard &  & \nolinkurl{kongming_lock/dual-panda/A14} \\
Lock-07 & Disassemble & Disassemble full lock & Hard & \checkmark & \nolinkurl{kongming_lock/dual-panda/D01} \\
Lock-08 & Disassemble & Disassemble key & Easy &  & \nolinkurl{kongming_lock/dual-panda/D06} \\
Lock-09 & Disassemble & Disassemble key + bar & Easy &  & \nolinkurl{kongming_lock/dual-panda/D11} \\
Lock-10 & Disassemble & Disassemble full lock & Hard &  & \nolinkurl{kongming_lock/dual-panda/D04} \\
Lock-11 & Disassemble & Disassemble key & Easy &  & \nolinkurl{kongming_lock/dual-panda/D09} \\
Lock-12 & Disassemble & Disassemble key + bar & Hard &  & \nolinkurl{kongming_lock/dual-panda/D14} \\
Lock-13 & Dex & Assemble full lock & Hard &  & \nolinkurl{kongming_lock/dual-floating-sharpa/A01} \\
Lock-14 & Dex & Assemble key & Easy & \checkmark & \nolinkurl{kongming_lock/dual-floating-sharpa/A06} \\
Lock-15 & Dex & Disassemble full lock & Hard &  & \nolinkurl{kongming_lock/dual-floating-sharpa/D01} \\
Lock-16 & Dex & Disassemble key & Easy &  & \nolinkurl{kongming_lock/dual-floating-sharpa/D06} \\
Cube-01 & Rubik & Solve scramble C01 & Easy &  & \nolinkurl{rubiks_panda/panda/C01} \\
Cube-02 & Rubik & Solve scramble C02 & Easy &  & \nolinkurl{rubiks_panda/panda/C02} \\
Cube-03 & Rubik & Solve scramble C03 & Easy &  & \nolinkurl{rubiks_panda/panda/C03} \\
Cube-04 & Rubik & Solve scramble C01 & Easy &  & \nolinkurl{rubiks_sharpa/floating-sharpa/C01} \\
Cube-05 & Rubik & Solve scramble C02 & Hard &  & \nolinkurl{rubiks_sharpa/floating-sharpa/C02} \\
Cube-06 & Rubik & Solve scramble C11 & Hard & \checkmark & \nolinkurl{rubiks_panda/panda/C11} \\
Cube-07 & Rubik & Solve scramble C12 & Hard &  & \nolinkurl{rubiks_panda/panda/C12} \\
Cube-08 & Rubik & Solve scramble C21 & Hard &  & \nolinkurl{rubiks_panda/panda/C21} \\
Cube-09 & Pocket & Clockwise top turn & Easy & \checkmark & \nolinkurl{pocket_cube/aloha/C01} \\
Cube-10 & Pocket & Clockwise top turn & Hard &  & \nolinkurl{pocket_cube/dual-sharpa/C01} \\
Cube-11 & Pocket & Half-turn in two stages & Easy &  & \nolinkurl{pocket_cube/aloha/C08} \\
Cube-12 & Pocket & Half-turn in two stages & Hard & \checkmark & \nolinkurl{pocket_cube/dual-sharpa/C08} \\
Home-01 & Tidy & Three upright books onto the middle shelf & Easy &  & \nolinkurl{bookshelf/dual-panda/BooksTableD1} \\
Home-02 & Tidy & Unpack a neatly packed crate & Easy &  & \nolinkurl{bookshelf/dual-panda/ShelfCrateD1} \\
Home-03 & Tidy & One book on the wrong layer of a tidy shelf & Hard & \checkmark & \nolinkurl{bookshelf/dual-panda/ShelfCrateFixD1} \\
Home-04 & Tidy & Books, a trophy and one box & Hard &  & \nolinkurl{bookshelf/dual-panda/ShelfTableD1} \\
Home-05 & Tidy & Three books and two pieces of trash & Hard &  & \nolinkurl{bookshelf/dual-panda/TableCleanD1} \\
Home-06 & Long & Prepare a shared writing desk & Easy & \checkmark & \nolinkurl{maniskill_long_horizon/fetch/L1} \\
Home-07 & Long & Prepare a tea station for two & Easy &  & \nolinkurl{maniskill_long_horizon/fetch/L3} \\
Home-08 & Long & Unload cups and cutlery from the dishwasher & Hard &  & \nolinkurl{maniskill_long_horizon/fetch/L5} \\
Home-09 & Long & Pack a picnic basket & Hard &  & \nolinkurl{maniskill_long_horizon/fetch/L6} \\
Home-10 & Precise & Insert three matching pegs & Hard &  & \nolinkurl{maniskill_geometry/panda/P1} \\
Home-11 & Precise & Seat a keyed electronics connector & Easy &  & \nolinkurl{maniskill_geometry/panda/P3} \\
Home-12 & Precise & Hang a mug by its handle & Hard & \checkmark & \nolinkurl{maniskill_geometry/panda/P5} \\
Home-13 & Precise & Insert a book into a narrow shelf gap & Easy &  & \nolinkurl{maniskill_geometry/panda/P7} \\
Home-14 & Dynamic & Stop a rolling orange & Easy & \checkmark & \nolinkurl{maniskill_dynamics/panda/D1} \\
Home-15 & Dynamic & Push a tall bottle without tipping & Hard &  & \nolinkurl{maniskill_dynamics/panda/D5} \\
Home-16 & Dynamic & Close a loaded drawer gently & Hard &  & \nolinkurl{maniskill_dynamics/panda/D6} \\
Home-17 & Dynamic & Slide a puck into a scoring zone & Easy &  & \nolinkurl{maniskill_dynamics/panda/D3} \\
\end{longtable}
}

\subsection{EP-Easy Index}
\label{app:epeasy}
EP-Easy contains the lower-scoring half of every task type. Table~\ref{tab:epeasy-index} lists its 50 tasks by type, using the suite index numbers of Table~\ref{tab:ep100-index}.

{\footnotesize
\begin{longtable}{@{}>{\raggedright\arraybackslash}p{0.2\linewidth}>{\raggedright\arraybackslash}p{0.08\linewidth}>{\raggedright\arraybackslash}p{0.66\linewidth}@{}}
\caption{EP-Easy index by task type.}\label{tab:epeasy-index}\\
\toprule
\textbf{Type} & \textbf{Tasks} & \textbf{Index numbers} \\
\midrule\endfirsthead
\toprule
\textbf{Type} & \textbf{Tasks} & \textbf{Index numbers} \\
\midrule\endhead
\bottomrule\endfoot
Launch-Aim & 4 & 02, 03, 04, 05 \\
Launch-Causal & 6 & 08, 09, 10, 11, 12, 15 \\
Launch-Glass & 3 & 21, 22, 23 \\
Launch-Rules & 4 & 26, 27, 28, 32 \\
Tower-Plan & 3 & 01, 02, 03 \\
Tower-Physics & 4 & 06, 07, 09, 11 \\
Tower-Dex & 4 & 14, 18, 20, 21 \\
Lock-Assemble & 3 & 02, 03, 05 \\
Lock-Disassemble & 3 & 08, 09, 11 \\
Lock-Dex & 2 & 14, 16 \\
Cube-Rubik & 4 & 01, 02, 03, 04 \\
Cube-Pocket & 2 & 09, 11 \\
Home-Tidy & 2 & 01, 02 \\
Home-Long & 2 & 06, 07 \\
Home-Precise & 2 & 11, 13 \\
Home-Dynamic & 2 & 14, 17 \\
\end{longtable}
}

\subsection{EP-Hard Index}
\label{app:ephard}
EP-Hard contains the higher-scoring half of every task type. Table~\ref{tab:ephard-index} lists its 50 tasks by type, using the suite index numbers of Table~\ref{tab:ep100-index}.

{\footnotesize
\begin{longtable}{@{}>{\raggedright\arraybackslash}p{0.2\linewidth}>{\raggedright\arraybackslash}p{0.08\linewidth}>{\raggedright\arraybackslash}p{0.66\linewidth}@{}}
\caption{EP-Hard index by task type.}\label{tab:ephard-index}\\
\toprule
\textbf{Type} & \textbf{Tasks} & \textbf{Index numbers} \\
\midrule\endfirsthead
\toprule
\textbf{Type} & \textbf{Tasks} & \textbf{Index numbers} \\
\midrule\endhead
\bottomrule\endfoot
Launch-Aim & 3 & 01, 06, 07 \\
Launch-Causal & 7 & 13, 14, 16, 17, 18, 19, 20 \\
Launch-Glass & 2 & 24, 25 \\
Launch-Rules & 5 & 29, 30, 31, 33, 34 \\
Tower-Plan & 2 & 04, 05 \\
Tower-Physics & 4 & 08, 10, 12, 13 \\
Tower-Dex & 4 & 15, 16, 17, 19 \\
Lock-Assemble & 3 & 01, 04, 06 \\
Lock-Disassemble & 3 & 07, 10, 12 \\
Lock-Dex & 2 & 13, 15 \\
Cube-Rubik & 4 & 05, 06, 07, 08 \\
Cube-Pocket & 2 & 10, 12 \\
Home-Tidy & 3 & 03, 04, 05 \\
Home-Long & 2 & 08, 09 \\
Home-Precise & 2 & 10, 12 \\
Home-Dynamic & 2 & 15, 16 \\
\end{longtable}
}

\subsection{EP-20 Index}
\label{app:ep20}
Launcher entries use loaded initialization only. EP-20 takes ten tasks from EP-Easy and ten from EP-Hard. Table~\ref{tab:ep20-index} lists them with their splits, embodiments, and native levels.

{\footnotesize
\setlength{\tabcolsep}{3pt}
\begin{longtable}{@{}>{\raggedright\arraybackslash}p{0.1\linewidth}>{\raggedright\arraybackslash}p{0.30\linewidth}>{\raggedright\arraybackslash}p{0.11\linewidth}>{\raggedright\arraybackslash}p{0.07\linewidth}>{\raggedright\arraybackslash}p{0.2\linewidth}>{\raggedright\arraybackslash}p{0.15\linewidth}@{}}
\caption{EP-20 index.}\label{tab:ep20-index}\\
\toprule
\textbf{Index} & \textbf{Task} & \textbf{Type} & \textbf{Split} & \textbf{Embodiment} & \textbf{Native level} \\
\midrule\endfirsthead
\toprule
\textbf{Index} & \textbf{Task} & \textbf{Type} & \textbf{Split} & \textbf{Embodiment} & \textbf{Native level} \\
\midrule\endhead
\bottomrule\endfoot
Launch-02 & Single-target warm-up & Aim & Easy & Dual Panda & tier 1 \\
Launch-08 & Orchard rolling bodies & Causal & Easy & Dual Panda & tier 2 \\
Launch-22 & Two glass layers & Glass & Easy & Dual Panda & tier 3 \\
Launch-19 & Movable flap at a narrow throat & Causal & Hard & Dual Panda & tier 4 \\
Launch-31 & Fragment-risk well & Rules & Hard & Dual Panda & tier 5 \\
Launch-34 & Material-ordered courtyard & Rules & Hard & Dual Panda & tier 5 \\
Tower-06 & Triangular transfer & Physics & Easy & Panda & L2, 3 moves \\
Tower-18 & Top-ring transfer & Dex & Easy & FR3--Sharpa & L1, 1 move \\
Tower-04 & Long five-ring rearrangement & Plan & Hard & Panda & L4, 31 moves \\
Tower-15 & Local two-ring transfer & Dex & Hard & Floating Sharpa & L2, 3 moves \\
Lock-03 & Assemble key + bar & Assemble & Easy & Dual Panda & key + bar, easy pose \\
Lock-14 & Assemble key & Dex & Easy & Dual floating Sharpa & key, easy pose \\
Lock-07 & Disassemble full lock & Disassemble & Hard & Dual Panda & full lock, easy pose \\
Cube-09 & Clockwise top turn & Pocket & Easy & ALOHA & 1 quarter turn \\
Cube-06 & Solve scramble C11 & Rubik & Hard & Panda & 14 quarter turns \\
Cube-12 & Half-turn in two stages & Pocket & Hard & Dual floating Sharpa & 2 quarter turns \\
Home-06 & Prepare a shared writing desk & Long & Easy & Fetch & 13 conditions \\
Home-14 & Stop a rolling orange & Dynamic & Easy & Panda & 3 conditions \\
Home-03 & One book on the wrong layer of a tidy shelf & Tidy & Hard & Dual Panda & D1 \\
Home-12 & Hang a mug by its handle & Precise & Hard & Panda & 4 conditions \\
\end{longtable}
}

\end{document}